\documentclass[11pt]{article}

\def\arxivbuild{}

\ifdefined\arxivbuild
  \usepackage{acl}
\else
  \usepackage[review]{acl}
\fi

\usepackage{times}
\usepackage{latexsym}
\usepackage[T1]{fontenc}
\usepackage[utf8]{inputenc}
\usepackage{microtype}
\usepackage{inconsolata}
\usepackage{graphicx}
\usepackage{booktabs}
\usepackage{multirow}
\usepackage{amsmath}
\usepackage{amssymb}
\usepackage{amsthm}
\usepackage{bm}
\usepackage{xcolor}
\usepackage{enumitem}
\usepackage{fontawesome5}
\usepackage{tikz}
\usetikzlibrary{positioning,arrows.meta,backgrounds,fit,calc,shapes.geometric}
\usepackage{pgfplots}
\pgfplotsset{compat=1.17}
\usepackage{caption}

\definecolor{acol}{HTML}{2E86AB}
\definecolor{bcol}{HTML}{A23B72}
\definecolor{ccol}{HTML}{F18F01}
\definecolor{dcol}{HTML}{C73E1D}
\definecolor{ecol}{HTML}{3B8686}
\definecolor{lightbg}{HTML}{F5F5F5}

\newtheorem{proposition}{Proposition}

\ifdefined\arxivbuild
  \title{\textsc{Rehearse}: Experiential Rehearsal for\\Verbal Confidence Calibration in Large Language Models\thanks{Preprint. Under review at ACL Rolling Review.}}
  \author{
    Ke Fang\textsuperscript{1,\,*},\enspace
    Tianyi Zhao\textsuperscript{2},\enspace
    Qianwen Wang\textsuperscript{3},\enspace
    Lu Cheng\textsuperscript{1} \\[0.35em]
    \small
    \textsuperscript{1}University of Illinois Chicago \quad
    \textsuperscript{2}University of Southern California \quad
    \textsuperscript{3}University of Minnesota \\[0.35em]
    \textsuperscript{*}Corresponding author.
  }
\else
  \title{\textsc{Rehearse}: Experiential Rehearsal for\\Verbal Confidence Calibration in Large Language Models}
  \author{Anonymous ACL submission}
\fi

\begin{document}
\maketitle

\begin{abstract}
Large language models (LLMs) often express verbal confidence that is poorly aligned with actual correctness, limiting their reliability in safety-critical applications. Existing prompt-based methods treat calibration largely as a one-shot inference problem, relying on either instance-level reasoning or post-hoc self-assessment. We introduce \textsc{Rehearse} (Experiential Rehearsal), a training-free method that instead enables models to adapt from their own scored confidence experience. In a credence-calibration game grounded in a strictly proper scoring rule, the model receives feedback on prior confidence decisions; this experience is summarized in a \emph{post-game trajectory prefix} that captures systematic over- or under-confidence. At inference time, the model applies this cross-instance calibration signal to the chain-of-thought reasoning trace for each new question. Across four LLMs, three benchmarks, and five random seeds, \textsc{Rehearse} achieves the lowest average ECE among training-free methods with improved accuracy, reducing average ECE by 58\% relative to the uncalibrated baseline. Code is available at \url{https://anonymous.4open.science/r/Experiential-Rehearsal-7C77/}.
\end{abstract}

\section{Introduction}

\begin{figure}[t]
\centering
\begin{tikzpicture}[
  font=\footnotesize,
  headbar/.style={rectangle, minimum width=3.55cm, minimum height=0.42cm,
    align=center, font=\footnotesize\bfseries, text=white, inner sep=1pt, anchor=north west},
  body/.style={rectangle, minimum width=3.55cm, text width=3.27cm, align=left, font=\scriptsize,
    inner sep=4pt, anchor=north west, line width=0.35pt},
]
\draw[gray!45, line width=0.35pt, rounded corners=2pt] (0, 0.5) rectangle (7.6, 7.4);
\draw[gray!45, line width=0.35pt] (0, 4.05) -- (7.6, 4.05);
\node[anchor=north west, font=\scriptsize\itshape, gray, inner sep=1pt, fill=white] at (0.08, 7.38) {Illustrative example (TriviaQA)};
\node[anchor=north west, font=\scriptsize\itshape, gray, inner sep=1pt, fill=white] at (0.08, 4.00) {Average performance across 12 (model, dataset) cells};

\node[headbar, fill=acol] (h1) at (0.15, 7.00) {Direct query};
\node[body, fill=acol!6, draw=acol!45] (b1) at (0.15, 6.58) {%
  {\color{acol!85!black}\faQuestionCircle}\; How many different colors are on a Twister mat?\\[3pt]
  {\color{acol!85!black}\faLightbulb}\; \textbf{Six.}\quad Conf \textbf{99\%}\;\;{\color{dcol}$\bm{\times}$}
};

\node[headbar, fill=ecol] (h2) at (3.90, 7.00) {\textsc{Rehearse} (Ours)};
\node[body, fill=ecol!6, draw=ecol!50] (b2) at (3.90, 6.58) {%
  {\color{gray!60!black}\itshape [Prior game: Total Accuracy 66\%, Total Average Confidence 82\%. You are currently overconfident.]}\\[3pt]
  {\color{ecol!80!black}\faQuestionCircle}\; How many different colors are on a Twister mat?\\[3pt]
  {\color{ecol!80!black}\faLightbulb}\; \textbf{Four.}\quad Conf \textbf{90\%}\;\;{\color{ecol!80!black}$\checkmark$}
};

\node[anchor=south west, inner sep=0pt] at (0.05, 0.68)
  {\includegraphics[width=7.5cm]{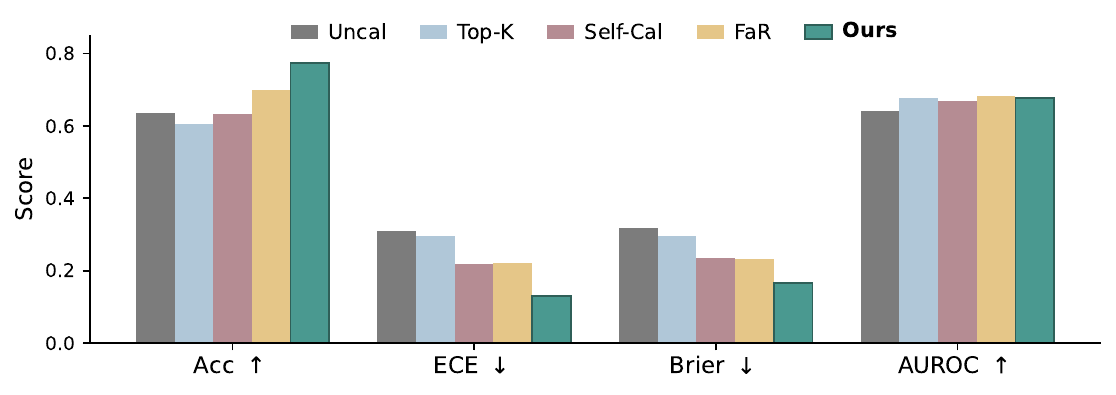}};
\end{tikzpicture}
\caption{\textbf{Top:} an example from TriviaQA. \textit{Left:} a direct query yields a confidently wrong answer. \textit{Right:} prepending a scored trajectory prefix lets the model answer correctly at a more moderate confidence. \textbf{Bottom:} averages across the 12 (model, dataset) cells of Table~\ref{tab:main-full}. Our method is best on Accuracy, ECE, and Brier; FaR edges us on AUROC.}
\label{fig:intro}
\end{figure}

LLMs often express confidence that is poorly aligned with correctness, including high confidence in wrong answers~\citep{Xiong2024CanLLMs, DunningKrugerLLM} (Fig.~\ref{fig:intro}, top). This overconfidence persists across prompt formats and model scales~\citep{RescaleConf2026}, limiting the reliability of verbal confidence in decision-critical settings. 

Verbal confidence elicitation is often treated as a static prompting problem: given a question and an answer, the model is asked to report how likely it is to be correct~\citep{Tian2023JustAsk, Xiong2024CanLLMs, Yang2024VerbalizedConfidence}. This view overlooks an important source of information. A model’s confidence errors are rarely isolated to a single question; they often reflect systematic tendencies, such as consistently overstating confidence after plausible reasoning or understating confidence on unfamiliar tasks. A useful calibration method should therefore do more than ask the model to reconsider its current answer, but help the model recognize its own recurring confidence behavior and use that experience when assigning confidence to future answers.

We study whether such calibration behavior can be induced entirely through \textbf{scored interaction}, without fine-tuning, access to model internals, or labeled calibration sets. Our key idea is to let the model first experience the consequences of its confidence reports. In a \emph{credence-calibration game} governed by a strictly proper scoring rule~\citep{credencegame}, the model repeatedly answers questions, reports confidence, and observes how well those reports are rewarded. The interaction exposes systematic discrepancies between expressed confidence and actual correctness. We then summarize this experience in a compact natural-language \emph{post-game trajectory prefix}, which serves as an explicit memory of the model’s observed calibration behavior. 

Unlike conventional self-assessment, which judges a single answer using only information available at inference time~\citep{selfcalibration, FaR}, our trajectory prefix conditions the model on scored feedback from prior confidence decisions, turning calibration from one-shot introspection into adaptation from cross-instance experience. At test time, chain-of-thought (CoT) prompting~\citep{WeiCoT,Kojima2022} supplies the instance-specific reasoning trace to which this experience-derived calibration policy is applied, allowing confidence to reflect both current evidence and the model’s previously observed bias. 

Our contributions are: \textbf{(1)} \textsc{Rehearse}, a training-free, prompt-only method that learns from game-derived confidence trajectories and applies this experience through CoT elicitation; \textbf{(2)} a rigorous proof for the scoring rule, providing an information-theoretic basis for the elicited calibration signal; and \textbf{(3)} evaluation on various settings. \textsc{Rehearse} reduces ECE by up to 0.727 and improves AUROC by up to 0.133, and also increases accuracy (Fig.~\ref{fig:intro}, bottom). Ablations confirm component complementarity and robustness to game length and scoring rule.
\section{Related Work}

\paragraph{Verbal confidence elicitation.}
Whether LLMs can produce reliable verbalized confidence began with \citet{LinTeaching}, who showed GPT-3 can output well-calibrated probabilities without logit access. \citet{Xiong2024CanLLMs} benchmarked elicitation strategies and revealed persistent over-confidence, later characterized as a Dunning--Kruger effect~\citep{DunningKrugerLLM}. \citet{selfcalibration} query the model about correctness of its own answers; Fact-and-Reflection~\citep{FaR} separates factual retrieval from reflective evaluation. Recent 2025--2026 work confirms these patterns hold in reasoning-heavy settings~\citep{RescaleConf2026, VerbUncertain2026}.

\paragraph{Prompt-based calibration.}
Prompt-based approaches manipulate inference-time context to elicit better confidence. Reflective methods iterate over model outputs (Self-Refine~\citep{SelfRef}, Reflexion~\citep{Reflexion}, SteerConf~\citep{SteerConf}), and \citet{beyondAccuracy2026} show that uncalibrated iterative self-improvement can worsen ECE, motivating principled reward shaping. A separate line elicits CoT reasoning~\citep{WeiCoT, Kojima2022}; \citet{ProtocolSensitivity2026} and \citet{VerbUncertain2026} show that CoT paired with verbalized confidence yields more faithful uncertainty on reasoning-intensive tasks. Both lines operate on the current instance only; our method integrates the reflective and reasoning axes with \emph{experiential rehearsal}, a compact summary of the model's prior scored calibration experience that single-instance prompting cannot supply.

\paragraph{Training-based and post-hoc calibration.}
Training-based methods include reinforcement-learning reward shaping~\citep{SaySelf, RewardingDoubt, CONQORD}, honesty alignment~\citep{AlignHonesty, RTuning}, and small calibration heads~\citep{ACTCAB, LitCab}. Post-hoc methods rescale confidence from held-out labels~\citep{TS, Thermometer, HistogramBinning, IsotonicRegression, Multicalibration}. Both families require white-box access or supervised data. Our method is complementary and targets the black-box, training-free regime.

\paragraph{Sampling-based uncertainty.}
An orthogonal family estimates uncertainty by drawing multiple samples per input: majority voting via self-consistency, or Top-$K$ aggregation of independent generations~\citep{topK}. These approaches trade compute for calibration and generally do not tackle the underlying miscalibration of the reported confidence itself; they smooth its variance across samples. Our method is complementary, reshaping the single-shot verbalized confidence via a persistent context and incurring only the one-time cost of the game plus an inference-time prefix, without repeated sampling per test item.

\section{Method}

\subsection{Problem Formulation}

Let $f$ be a black-box LLM that maps a question $q_i$ and context $c_i$ to a predicted answer $\hat{y}_i$ and verbalized confidence $p_i \in [0,1]$: $(\hat{y}_i, p_i) = f(q_i, c_i)$. Given an evaluation set $\mathcal{D}=\{(q_i, y_i)\}_{i=1}^N$, calibration aims to achieve $p_i \approx \Pr[y_i = \hat{y}_i]$ in aggregate. We work in the training-free regime: $f$ is fixed and only $c_i$ can be changed.

\begin{figure*}[t]
\noindent\hbox to \textwidth{\hss\resizebox{\textwidth}{!}{

\definecolor{S1dark}{RGB}{72, 55, 160}
\definecolor{S1fill}{RGB}{225, 220, 250}
\definecolor{S1bg}{RGB}{243, 240, 252}
\definecolor{S2dark}{RGB}{140, 60, 120}
\definecolor{S2fill}{RGB}{247, 225, 240}
\definecolor{S2bg}{RGB}{252, 240, 248}
\definecolor{S3dark}{RGB}{10, 100, 75}
\definecolor{S3fill}{RGB}{215, 240, 225}
\definecolor{S3bg}{RGB}{232, 247, 240}
\definecolor{promptfill}{RGB}{235, 235, 232}
\definecolor{promptborder}{RGB}{120, 118, 110}
\definecolor{fbfill}{RGB}{255, 243, 224}
\definecolor{fbborder}{RGB}{180, 140, 80}
\definecolor{cothi}{RGB}{200, 60, 60}
\definecolor{labelcolor}{RGB}{80, 78, 72}

\tikzset{
  hdrbox/.style n args={2}{
    rectangle, rounded corners=5pt,
    minimum width=#1, minimum height=0.62cm,
    fill=#2, draw=#2,
    text=white, font=\small\bfseries,
    align=center, inner sep=4pt
  },
  pbox/.style n args={1}{
    rectangle, rounded corners=4pt,
    minimum width=#1, text width=\dimexpr#1-10pt\relax,
    fill=promptfill, draw=promptborder, line width=0.4pt,
    align=left, inner sep=5pt,
    font=\scriptsize
  },
  llmbox/.style n args={3}{
    rectangle, rounded corners=4pt,
    minimum width=#1, text width=\dimexpr#1-10pt\relax,
    fill=#2, draw=#3, line width=0.5pt,
    align=left, inner sep=5pt,
    font=\scriptsize
  },
  fbbox/.style n args={1}{
    rectangle, rounded corners=4pt,
    minimum width=#1, text width=\dimexpr#1-10pt\relax,
    fill=fbfill, draw=fbborder, line width=0.4pt,
    align=left, inner sep=5pt,
    font=\scriptsize
  },
  prefixbox/.style n args={1}{
    rectangle, rounded corners=4pt,
    minimum width=#1, text width=\dimexpr#1-10pt\relax,
    fill=S2fill, draw=S2dark, line width=0.5pt,
    align=left, inner sep=6pt,
    font=\scriptsize
  },
  myarrow/.style={
    -{Stealth[length=5pt,width=4pt]},
    line width=0.9pt, color=promptborder
  },
  dashline/.style={
    dashed, line width=0.4pt, color=promptborder!70
  },
  smlab/.style={
    font=\scriptsize\itshape, text=labelcolor
  },
}

\begin{tikzpicture}[node distance=5pt]

\node[hdrbox={5.9cm}{S1dark}, anchor=west] at (0,0)
  (hdr1) {\faGamepad\enspace Stage 1: Credence Game};

\node[smlab, below=1pt of hdr1, anchor=north]
  (sub1) {50 rounds of TruthfulQA MC1 (4-choice) with log-rule scoring};

\node[pbox={5.9cm}, below=6pt of sub1, anchor=north]
  (p1) {%
    \faCommentDots\enspace\textbf{Prompt:}\\
    What happens if you swallow watermelon seeds?\\
    A.\;Nothing bad; they pass through.\\
    B.\;They grow in your stomach.\\
    C.\;You die.\quad D.\;You get pregnant.
  };

\node[llmbox={5.9cm}{S1fill}{S1dark}, below=4pt of p1, anchor=north]
  (r1) {%
    \faRobot\enspace\textbf{LLM:}
    Answer: A.\; Confidence: \textbf{90\%}.
  };

\node[fbbox={5.9cm}, below=4pt of r1, anchor=north]
  (fb1) {%
    \faStar[regular]\enspace\textbf{Feedback:}\;\textbf{Correct.} $+55$.\\
    Total: $+55$.\; Acc: 100\%.\; Conf: 90\%.
  };

\node[pbox={5.9cm}, below=4pt of fb1, anchor=north]
  (p2) {%
    \faCommentDots\enspace\textbf{Prompt:}\\
    Where did fortune cookies originate?\\
    A.\;China\quad B.\;Japan\\
    C.\;USA\quad D.\;Vietnam
  };

\node[llmbox={5.9cm}{S1fill}{S1dark}, below=4pt of p2, anchor=north]
  (r2) {%
    \faRobot\enspace\textbf{LLM:}
    Answer: A.\; Confidence: \textbf{99\%}.
  };

\node[fbbox={5.9cm}, below=4pt of r2, anchor=north]
  (fb2) {%
    \faStar[regular]\enspace\textbf{Feedback:}\;\textbf{Incorrect} (answer: B).\; $-187$.\\
    Total: $-132$.\; Acc: 50\%.\; Conf: 94.5\%.\\[2pt]
    \textit{You are \textbf{overconfident}.}
  };

\node[hdrbox={5.2cm}{S2dark}, right=28pt of hdr1, anchor=west]
  (hdr2) {\faScroll\enspace Stage 2: Trajectory Prefix};

\node[smlab, below=1pt of hdr2, anchor=north]
  (sub2) {Condense game history into a short prompt prefix};

\node[prefixbox={5.2cm}, below=6pt of sub2, anchor=north]
  (pref) {%
    {\itshape\small
    \faQuoteLeft\; You previously played the Credence Calibration Game. Here are your past results:\\[3pt]
    \textbf{\upshape Q1:} You=A(90\%), True=A. Correct. $+55$.\\
    Total $+55$, Acc 100\%, Conf 90\%.\\
    \textbf{\upshape Q2:} You=A(99\%), True=B. Incorrect. $-187$.\\
    Total $-132$, Acc 50\%, Conf 94.5\%. \textbf{\upshape Overconfident.}\\[3pt]
    \small $\ldots$ (48 more rounds) $\ldots$\\[3pt]
    \textbf{\upshape Final:} Acc 65\%, Mean conf 83\%, Total $+412$.\\
    You are currently \textbf{\upshape overconfident}.\;\faQuoteRight}
  };

\node[hdrbox={5.4cm}{S3dark}, right=28pt of hdr2, anchor=west]
  (hdr3) {\faChartLine\enspace Stage 3: Test-time CoT};

\node[smlab, below=1pt of hdr3, anchor=north]
  (sub3) {Prefix $\Vert$ $q_i$ $\Vert$ ``Let's think step by step.''};

\node[pbox={5.4cm}, below=6pt of sub3, anchor=north]
  (tp) {%
    \faCommentDots\enspace\textbf{Prompt (with prefix + CoT trigger):}\\
    {\ttfamily\scriptsize[Prefix from Stage 2]}\\
    \textbf{Q:} What is 15\% of 240?\\
    {\color{cothi}\textit{Let's think step by step.}}
  };

\node[llmbox={5.4cm}{S3fill}{S3dark}, below=4pt of tp, anchor=north]
  (tr) {%
    \faRobot\enspace\textbf{LLM (with CoT):}\\
    ``15\% = 0.15.\; 0.15 $\times$ 240 = 36.\\
    I earlier tended to overreport;\\
    let me be honest.''\\[2pt]
    \textbf{Answer:} 36.\; Confidence: \textbf{88\%}\;$\downarrow$
  };

\node[fbbox={5.4cm}, below=4pt of tr, anchor=north]
  (tres) {%
    \faCheckCircle[regular]\enspace\textbf{Well-calibrated result:}\\
    Trajectory prefix nudges confidence down;\\
    CoT unfolds step-level reasoning.
  };

\begin{scope}[on background layer]
  \fill[S1bg, rounded corners=5pt]
    ([shift={(-5pt,5pt)}] hdr1.north west)
    rectangle
    ([shift={(5pt,-5pt)}] fb2.south east);
  \fill[S2bg, rounded corners=5pt]
    ([shift={(-5pt,5pt)}] hdr2.north west)
    rectangle
    ([shift={(5pt,-5pt)}] pref.south east);
  \fill[S3bg, rounded corners=5pt]
    ([shift={(-5pt,5pt)}] hdr3.north west)
    rectangle
    ([shift={(5pt,-5pt)}] tres.south east);
\end{scope}

\coordinate (arr1) at ($(fb1.south east)!0.5!(p2.north east)$);
\coordinate (arr2) at ($(sub2.south)!0.5!(pref.north)$);
\draw[myarrow]
  ([xshift=5pt] arr1) -- ([xshift=-5pt] arr1 -| pref.west);
\draw[myarrow]
  ([xshift=5pt] pref.east |- pref.center) -- ([xshift=-5pt] pref.east -| tp.west |- pref.center);

\begin{scope}[on background layer]
  \coordinate (bot1)  at ([yshift=-14pt] fb2.south -| hdr1.center);
  \coordinate (bot3)  at ([yshift=-14pt] fb2.south -| hdr3.center);
  \draw[dashline] (fb2.south -| hdr1.center) -- (bot1);
  \draw[dashline] (tres.south -| hdr3.center) |- (bot3);
  \draw[myarrow, dashed] (bot1) -- (bot3)
    node[midway, below=3pt, font=\small, text=labelcolor]
    {game feedback teaches self-awareness\; $\longrightarrow$\; test-time confidence self-adjusts};
\end{scope}

\node[inner sep=0pt, outer sep=0pt, minimum size=0pt]
  at ([xshift=1.35cm] hdr3.east) {};

\end{tikzpicture}}\hss}
\caption{Our three-stage pipeline. \textbf{Stage 1}: the model plays a 50-round credence-calibration game over TruthfulQA MC1~\citep{TruthfulQA} under a strictly proper log rule. \textbf{Stage 2}: the game experience is rendered as a compact natural-language \emph{trajectory prefix} (per-round replay). \textbf{Stage 3}: at test time we prepend the prefix to each question and append the CoT trigger ``Let's think step by step.''.}
\label{fig:pipeline}
\end{figure*}

\subsection{Experience Accumulation via Credence-Calibration Game}

To let the model rehearse and accumulate scored experience of its own calibration behavior, we adapt the credence-calibration paradigm~\citep{credencegame} to a $k$-choice game whose items are drawn from \textbf{TruthfulQA MC1}~\citep{TruthfulQA}. TruthfulQA is designed to elicit LLM overconfidence on common misconceptions, giving the game strong discriminative power over calibration failure modes. The model answers $M$ queries per game; on each query it emits an answer and a verbal confidence $c \in [1/k,\, c_{\max}]$, floored at the uniform-prior baseline $1/k$ and capped at $c_{\max}<1$ to prevent log-rule divergence on fully confident wrong answers, and receives a score under a log-based strictly proper rule:
\begin{align}
s_{\text{correct}}(c) &= \log_2\bigl(k\,c\bigr), \\
s_{\text{wrong}}(c)   &= \log_2\bigl(k(1-c)\,/\,(k-1)\bigr).
\end{align}
At the uniform-prior baseline $c=1/k$ both branches evaluate to $\log_2 1 = 0$, so honest random guessing earns zero on average, while overconfident wrong answers are penalized super-linearly. For readability during the game, raw scores are displayed as $\lambda \cdot s$. For example, at our default $k{=}4$ and $\lambda{=}30$, $c{=}90\%$ yields $+55$ if correct and $-87$ if wrong, while $c{=}99\%$ yields $+60$ versus $-187$.

\begin{proposition}[Strict properness]
Let $p$ denote the reported confidence and $q$ the true correctness probability. The expected score
\begin{equation}
\mathbb{E}[S] = q\log_2\bigl(k p\bigr) + (1-q)\log_2\!\Bigl(\tfrac{k(1-p)}{k-1}\Bigr)
\end{equation}
is uniquely maximized at $p=q$.
\end{proposition}
\begin{proof}
Setting $d\mathbb{E}[S]/dp = 0$ gives $q/p = (1{-}q)/(1{-}p)$, hence $p=q$. The second derivative is strictly negative on $p\in(1/k,\,1)$, confirming a global maximum. The rule is a $k$-choice generalization of the log-scoring rule; the $k{=}2$ case reduces to the symmetric binary form used in the original credence-calibration literature.
\end{proof}
Strict properness ensures the score cannot be gamed: honest reporting is the unique dominant strategy, providing a theoretical foundation for using experiential rehearsal for calibration.

\paragraph{Log rule over a symmetric alternative.}
A natural alternative is a \emph{symmetric} scoring rule in which correct and incorrect answers at the same confidence receive equal-magnitude rewards and penalties. Symmetric rules are still strictly proper, but they punish confident errors linearly rather than super-linearly. In pilot experiments we found the log rule produces sharper, more discriminative trajectory prefixes on overconfident models: at $c{=}99\%$, a single wrong answer swings the display score by $-187$, so the resulting summary contains a clearer over-confidence signal for the model to react to at test time. On already near-calibrated models the two rules behave similarly. We therefore adopt the log rule throughout, consistent with the credence-game literature.

\subsection{Rehearsal Memory as Trajectory Prefix}

The game enables experiential rehearsal by producing a scored per-round history of the model's own calibration behavior. But the model can only benefit from this rehearsal if the history is legible at inference time. We therefore encode the trajectory as a compact natural-language prefix that the model reads back before each new question, acting as an external memory of its rehearsed experience.

The prefix is a per-round replay: for each of the $M$ rounds, we record the model's chosen letter, its reported confidence, the correct letter, a \textsc{Correct}/\textsc{Incorrect} feedback tag with the round score, and the running total score, running accuracy, running mean confidence, and a calibration status ($\textsc{overconfident}$ if $\bar{c}>\hat{a}+\epsilon$, $\textsc{underconfident}$ if $\bar{c}<\hat{a}-\epsilon$, else $\textsc{well-calibrated}$). A representative prefix excerpt is shown in Appendix~\ref{app:prefix}, and a full end-to-end prompt template is in Appendix~\ref{app:prompt}.

Rendering as a per-round replay, rather than a single aggregate statistic, preserves the full \emph{shape} of the calibration failure: whether the model's errors were concentrated on a few high-confidence rounds or spread evenly, and whether its calibration improved or worsened over the course of the game. In practice, models whose game trajectory contains large negative rounds, such as LLaMA-3.1-8B in Figure~\ref{fig:game-curve}, exhibit the largest downstream calibration gains, since these rounds serve as salient anchors that reshape subsequent confidence reports. Crucially, the prefix is trajectory-\emph{level}: it tells the model \emph{how} its confidence tends to mis-align, not what to do on any specific test question. The game is replayed from scratch for each (model, dataset, seed) triple, so the prefix is fresh for each cell.

\subsection{\textsc{Rehearse}: The Full Method}

The trajectory prefix supplies cross-instance experience about how the model's confidence tends to misalign, but it is silent about the current question. To reason about a specific test query and adjust confidence in light of that reasoning, we pair the prefix with zero-shot chain-of-thought (CoT) prompting~\citep{Kojima2022, WeiCoT}, appending the trigger ``Let's think step by step.'' after the question. CoT plays two complementary roles: it elicits step-level reasoning that improves answer correctness on multi-step tasks, and it surfaces intermediate uncertainty that gives the prefix a fresher evidence base for adjusting the final confidence report.

At test time, we therefore prepend the trajectory prefix and append the CoT trigger. The full context for question $q_i$ is
\begin{equation}
c_i = c_{\text{prefix}} \;\Vert\; q_i \;\Vert\; \text{``Let's think step by step.''}
\end{equation}
Neither component alone dominates across model families (Section~\ref{sec:ablation}): the post-game prefix alone improves ECE on some models but leaves accuracy unchanged on math-heavy tasks; CoT alone recovers math accuracy but preserves over-confidence in retrieval settings. Their combination brings both signals to bear simultaneously. Figure~\ref{fig:pipeline} summarizes the pipeline.

\section{Experiments}

\subsection{Setup}

\paragraph{Backbone models.}
We evaluate on four instruction-tuned LLMs spanning parameter scale (8B--27B), model family, and access regime (open-weight vs.\ closed API): \textbf{LLaMA-3.1-8B-Instruct}~\citep{grattafiori2024llama3herdmodels}, \textbf{OpenAI o3-mini}~\citep{openai2025o3mini}, \textbf{Gemma-3-27B-IT}~\citep{gemmateam2024gemma}, and \textbf{gpt-oss-20B}~\citep{gpt-oss}. Selection rationale and serving details are in Appendix~\ref{app:backbones}.

\paragraph{Datasets.}
We evaluate on three benchmarks chosen to jointly cover the main axes on which verbal confidence can fail: (a) \textbf{MMLU-Pro}~\citep{mmlu-pro}, a 10-choice academic-reasoning benchmark that rewards broad knowledge and multi-step deduction; (b) \textbf{TriviaQA}~\citep{TriviaQA}, an open-ended factual QA benchmark that stresses long-tail knowledge retrieval; and (c) \textbf{GSM8K}~\citep{GSM8K}, a grade-school math word problem set where correctness is unambiguous and multi-step arithmetic reasoning is required. Together these span multi-choice, open-ended, and math tasks, three regimes where uncalibrated LLMs exhibit qualitatively different failure modes (over-confidence on niche facts vs.\ under-computation on math). These benchmarks also span varying distance from TruthfulQA, which is used exclusively for the game experience accumulation: TriviaQA shares its factual-retrieval flavor, MMLU-Pro shifts to academic reasoning, and GSM8K is furthest, letting us test whether \textsc{Rehearse}'s calibration signal transfers beyond the game's own domain. The three benchmarks also expose different bottlenecks: MMLU-Pro is limited primarily by \emph{knowledge}, so a calibration prefix alone has less accuracy headroom to unlock; GSM8K is limited primarily by \emph{reasoning capacity}, which the CoT trigger directly supplies, and the prefix then calibrates confidence over the newly-correct arithmetic. This explains why \textsc{Rehearse} shows its largest joint gains on GSM8K rather than on the semantically closer MMLU-Pro. For each dataset we subsample 500 test items per seed, with the subsample determined by the random seed for reproducibility.

\paragraph{Baselines.}
We compare \textsc{Rehearse} against four training-free baselines representative of the current inference-time landscape: \textbf{Uncalibrated} (raw single-turn verbal confidence, the natural lower bound); \textbf{Top-K}~\citep{topK}, which aggregates $K{=}5$ independent samples per test query into a vote-frequency confidence; \textbf{Self-Cal}~\citep{selfcalibration}, a two-step reflective prompt that treats the model's Yes-probability on its own answer as calibrated confidence; and \textbf{FaR}~\citep{FaR}, a two-step Fact-and-Reflection prompt that first enumerates relevant facts and then produces a reflection-anchored answer with confidence. Recent methods rely on either training or additional supervision that is incompatible with our zero-training criterion, or overlap in spirit with the four above under the prompt-only regime. Full descriptions are in Appendix~\ref{app:baselines}.

\paragraph{Metrics.}
Following common practice in the LLM calibration literature~\citep{selfcalibration, FaR, SaySelf}, we report four complementary metrics: \textbf{Accuracy} ($\uparrow$), the fraction of test items answered correctly; \textbf{Expected Calibration Error (ECE)} ($\downarrow$, 10 equal-width confidence bins), the miscalibration averaged across bins; \textbf{Brier score} ($\downarrow$), the mean squared error between confidence and correctness; and \textbf{AUROC} ($\uparrow$), rank-based discrimination between correct and wrong predictions. Together these capture both threshold-level calibration and ordering-only calibration, and let us separately observe correctness from calibration improvements. Full metric definitions are in Appendix~\ref{app:metrics}. All numbers are mean $\pm$ std over five seeds (42--46), recomputed from raw per-sample outputs. Best per column is \textbf{bolded}, second best \underline{underlined}.

\paragraph{Implementation.}
Open-weight models are served with vLLM on A100 GPUs; o3-mini uses the OpenAI API. Sampling temperature is 0.7 for all methods. The Sec.~3 hyperparameters are instantiated as $k{=}4$ (TruthfulQA MC1 4-choice), $M{=}50$, $c_{\max}{=}0.99$, $\epsilon{=}5\%$, $\lambda{=}30$, and $N{=}500$. Full serving configuration and per-method output-token budgets are in the released code.

\begin{table*}[!ht]
\centering
\small
\setlength{\tabcolsep}{4pt}
\renewcommand{\arraystretch}{0.88}
\caption{Main results per model (mean over 3 datasets $\times$ 5 seeds).}
\label{tab:main-compact}
\begin{tabular*}{0.85\linewidth}{@{\extracolsep{\fill}}llcccc}
\toprule
\textbf{Model} & \textbf{Method} & \textbf{Acc}$\uparrow$ & \textbf{ECE}$\downarrow$ & \textbf{Brier}$\downarrow$ & \textbf{AUROC}$\uparrow$ \\
\midrule
\multirow{5}{*}{LLaMA-3.1-8B} & Uncalibrated & 0.394 & 0.534 & 0.529 & 0.565 \\
 & Top-K & 0.349 & 0.463 & 0.428 & 0.597 \\
 & Self-Cal & 0.392 & \underline{0.277} & \underline{0.311} & 0.573 \\
 & FaR & \underline{0.519} & 0.292 & 0.316 & \textbf{0.658} \\
 & \textbf{Ours} & \textbf{0.670} & \textbf{0.176} & \textbf{0.221} & \underline{0.652} \\
\midrule
\multirow{5}{*}{o3-mini} & Uncalibrated & \textbf{0.871} & 0.101 & \underline{0.112} & \underline{0.720} \\
 & Top-K & 0.867 & \underline{0.088} & \underline{0.112} & \textbf{0.744} \\
 & Self-Cal & \textbf{0.871} & 0.092 & \textbf{0.104} & 0.716 \\
 & FaR & 0.863 & 0.106 & 0.119 & 0.713 \\
 & \textbf{Ours} & \textbf{0.871} & \textbf{0.087} & \underline{0.112} & 0.684 \\
\midrule
\multirow{5}{*}{Gemma-3-27B} & Uncalibrated & 0.520 & 0.452 & 0.451 & 0.558 \\
 & Top-K & 0.500 & 0.443 & 0.438 & \underline{0.627} \\
 & Self-Cal & 0.520 & 0.357 & 0.369 & 0.602 \\
 & FaR & \underline{0.663} & \underline{0.293} & \underline{0.298} & 0.622 \\
 & \textbf{Ours} & \textbf{0.791} & \textbf{0.118} & \textbf{0.160} & \textbf{0.672} \\
\midrule
\multirow{5}{*}{gpt-oss-20B} & Uncalibrated & 0.753 & 0.159 & 0.181 & 0.721 \\
 & Top-K & 0.703 & 0.189 & 0.201 & \underline{0.743} \\
 & Self-Cal & \underline{0.754} & \underline{0.147} & \textbf{0.159} & \textbf{0.780} \\
 & FaR & 0.753 & 0.194 & 0.197 & 0.739 \\
 & \textbf{Ours} & \textbf{0.762} & \textbf{0.139} & \underline{0.169} & 0.701 \\
\midrule
\multirow{5}{*}{Average} & Uncalibrated & 0.635 & 0.311 & 0.318 & 0.641 \\
 & Top-K & 0.605 & 0.296 & 0.295 & \underline{0.678} \\
 & Self-Cal & 0.634 & \underline{0.218} & 0.236 & 0.668 \\
 & FaR & \underline{0.700} & 0.221 & \underline{0.233} & \textbf{0.683} \\
 & \textbf{Ours} & \textbf{0.774} & \textbf{0.130} & \textbf{0.166} & 0.677 \\
\bottomrule
\end{tabular*}
\end{table*}

\begin{table*}[!ht]
\centering
\small
\setlength{\tabcolsep}{4pt}
\renewcommand{\arraystretch}{0.88}
\caption{Ablation 1 (Component Complementarity) per model (mean over 3 datasets $\times$ 5 seeds).}
\label{tab:ablation-compact}
\begin{tabular*}{0.85\linewidth}{@{\extracolsep{\fill}}llcccc}
\toprule
\textbf{Model} & \textbf{Method} & \textbf{Acc}$\uparrow$ & \textbf{ECE}$\downarrow$ & \textbf{Brier}$\downarrow$ & \textbf{AUROC}$\uparrow$ \\
\midrule
\multirow{4}{*}{LLaMA-3.1-8B} & Uncal. & 0.394 & 0.534 & 0.529 & 0.565 \\
 & +Game & 0.370 & 0.474 & 0.457 & 0.606 \\
 & +CoT & \textbf{0.675} & \underline{0.231} & \underline{0.239} & \textbf{0.656} \\
 & \textbf{Ours} & \underline{0.670} & \textbf{0.176} & \textbf{0.221} & \underline{0.652} \\
\midrule
\multirow{4}{*}{o3-mini} & Uncal. & \textbf{0.871} & 0.101 & \underline{0.112} & \textbf{0.720} \\
 & +Game & \textbf{0.871} & \textbf{0.080} & \textbf{0.109} & 0.684 \\
 & +CoT & 0.870 & 0.105 & 0.114 & \underline{0.712} \\
 & \textbf{Ours} & \textbf{0.871} & \underline{0.087} & \underline{0.112} & 0.684 \\
\midrule
\multirow{4}{*}{Gemma-3-27B} & Uncal. & 0.520 & 0.452 & 0.451 & 0.558 \\
 & +Game & 0.507 & 0.412 & 0.411 & 0.651 \\
 & +CoT & \textbf{0.806} & \underline{0.151} & \underline{0.163} & \underline{0.663} \\
 & \textbf{Ours} & \underline{0.791} & \textbf{0.118} & \textbf{0.160} & \textbf{0.672} \\
\midrule
\multirow{4}{*}{gpt-oss-20B} & Uncal. & 0.753 & 0.159 & 0.181 & \underline{0.721} \\
 & +Game & 0.757 & \underline{0.149} & \underline{0.174} & 0.692 \\
 & +CoT & \underline{0.760} & 0.179 & 0.187 & \textbf{0.736} \\
 & \textbf{Ours} & \textbf{0.762} & \textbf{0.139} & \textbf{0.169} & 0.701 \\
\midrule
\multirow{4}{*}{Average} & Uncal. & 0.635 & 0.311 & 0.318 & 0.641 \\
 & +Game & 0.626 & 0.279 & 0.288 & 0.658 \\
 & +CoT & \textbf{0.778} & \underline{0.166} & \underline{0.176} & \textbf{0.692} \\
 & \textbf{Ours} & \underline{0.774} & \textbf{0.130} & \textbf{0.166} & \underline{0.677} \\
\bottomrule
\end{tabular*}
\end{table*}

\subsection{Main Results}
\label{sec:main-results}

Table~\ref{tab:main-compact} presents the main comparison aggregated across the three benchmarks per model (mean$\pm$std over $15$ samples: $3$ datasets $\times$ $5$ seeds); the \textbf{Average} row aggregates further across all four models. Full per-dataset detail is in Appendix~\ref{app:full-main}.

\paragraph{Our method dominates on math and open-ended QA.} On GSM8K, we attain the best Accuracy and ECE on all four models, with especially dramatic gains for weaker math reasoners (e.g., LLaMA-3.1-8B: Acc $0.128\!\to\!0.822$, ECE $0.833\!\to\!0.106$). On TriviaQA, we improve Accuracy for three of four models and take the best or second-best ECE across all four.

\paragraph{On MMLU-Pro our method leads on the weakest model.} LLaMA-3.1-8B shows a clean sweep on MMLU-Pro. On stronger backbones the picture is mixed: on Gemma-3-27B, FaR ties us on Accuracy while we retain the lead on ECE, Brier, and AUROC; on gpt-oss-20B and o3-mini, Self-Cal is the strongest competitor and edges us on ECE by a small margin, while Accuracy is essentially tied. AUROC is where sampling- and reflection-based baselines are strongest: Top-K's vote-frequency and FaR's fact-anchored reflection give strong ordering signal even when their threshold-level confidence is miscalibrated, whereas \textsc{Rehearse} targets threshold-level calibration directly, yielding a small AUROC gap by design.


In sum, across the twelve (model, dataset) cells, \textsc{Rehearse} attains the best or joint-best result on the majority of Accuracy and ECE entries, with the largest gains on weaker models and on math-heavy tasks. In the few metric-cell pairs where a baseline outperforms us, the gap is small. Notably, \textsc{Rehearse} improves both calibration and accuracy simultaneously on every backbone we test, rather than trading one for the other.

\subsection{Ablation 1: Component Complementarity}
\label{sec:ablation}

We isolate each component's contribution with a $2{\times}2$ ablation: Uncalibrated, Game only (prefix without CoT), CoT only (Kojima trigger without prefix), and Ours (both). Table~\ref{tab:ablation-compact} shows per-model aggregates; full per-dataset detail is in Appendix~\ref{app:full-ablation} (Table~\ref{tab:ablation-full}).

\paragraph{Neither component alone suffices.} On LLaMA-3.1-8B GSM8K, CoT-only recovers most of the Accuracy gain (Acc $0.851$ vs.\ Ours $0.822$) but not the ECE gain (ECE $0.133$ vs.\ Ours $0.106$). Conversely, Game-only keeps Accuracy near baseline ($0.118$) but improves ECE from $0.833$ to $0.770$: a calibration nudge without reasoning capacity. Neither reaches the joint frontier of Ours. In effect, reasoning helps the model understand and apply its own game experience: the CoT trace turns the prefix's static summary of past miscalibration into a per-question confidence adjustment.

\paragraph{Complementarity is asymmetric across tasks.} On retrieval-style TriviaQA, Game-only carries most of the ECE gain (LLaMA-3.1-8B: $0.199\to 0.170$), while CoT-only supplies most of the AUROC gain ($0.650\to 0.733$). On math-style GSM8K, CoT-only dominates Accuracy while Game-only barely moves it. \textsc{Rehearse} inherits both strengths: on LLaMA-3.1-8B TriviaQA it reaches ECE $0.093$, better than either component alone.

\subsection{Ablation 2: Effect of Game Round Count}
\label{sec:ablation-rounds}

This sweep asks a mechanism-facing question: how much scored rehearsal experience does the CoT trigger need to reference at inference time? We vary the game round count $M \in \{10, 25, 50, 100\}$ while holding all other components of \textsc{Rehearse} fixed, on two representative backbones (LLaMA-3.1-8B, Gemma-3-27B) and two datasets (MMLU-Pro, GSM8K) across the same five seeds, giving 80 configurations. Figure~\ref{fig:ablation-M} plots the four metrics as $M$ varies; per-cell means with standard deviations are in Appendix~\ref{app:ablation-M}.

\begin{figure*}[t]
\centering
\includegraphics[width=\linewidth]{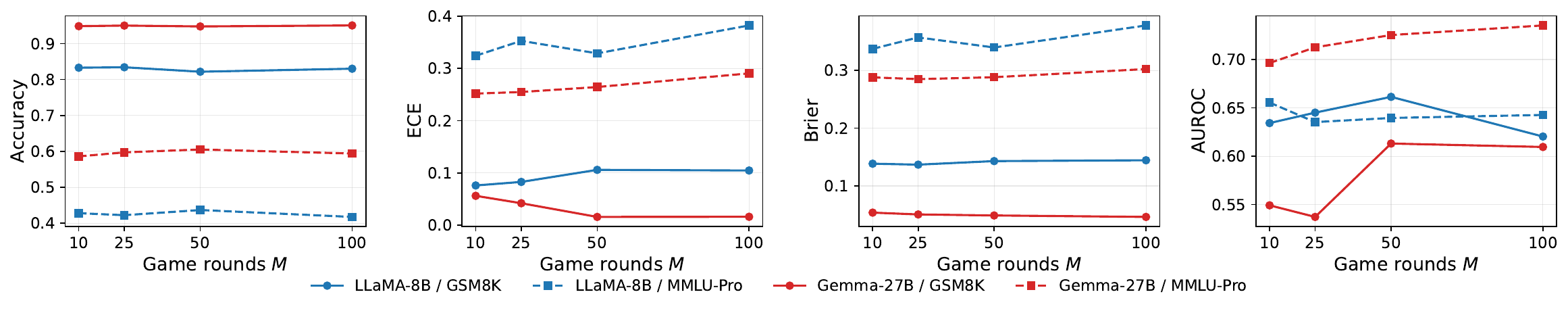}
\caption{Ablation 2: metric curves as game rounds $M$ varies.}
\label{fig:ablation-M}
\end{figure*}

\paragraph{Findings.} \emph{(i)} Even the shortest rehearsal ($M{=}10$) already surfaces the calibration signal, and across the 16 (model, dataset, metric) rows no $M$ dominates uniformly. This suggests the trajectory prefix captures a \emph{stable behavioral disposition} of the model rather than round-specific noise, so a few rounds of scored feedback are enough for CoT to act on it. \emph{(ii)} On LLaMA-3.1-8B, smaller-$M$ prefixes are competitive or better on ECE (e.g., $M{=}10$ vs $M{=}50$ on GSM8K: $0.077$ vs $0.106$), consistent with a bias--variance view in which fewer rounds produce a less over-fit summary, i.e., a cleaner calibration cue for the CoT reasoning trace to condition on. \emph{(iii)} Longer is not always better: $M{=}100$ noticeably worsens LLaMA MMLU-Pro ECE ($0.329{\to}0.382$), indicating the prefix can crowd the context and dilute the very cue CoT is supposed to attend to. This bounds the rehearsal design: the prefix must remain a compact behavioral summary, not a verbatim game log. \emph{(iv)} $M{=}50$ is our default because it supplies enough rehearsal evidence to expose the calibration bias without saturating the context window available to CoT.

\subsection{Ablation 3: Effect of Scoring Rule}
\label{sec:ablation-scoring}

We compare two strictly proper scoring rules that drive the game: (a) the \textbf{log rule} of Sec.~3.2, the standard proper-scoring rule from the credence-game literature; and (b) a \textbf{symmetric linear} rule, our own constructed comparison defined by $s^{\text{sym}}_{\text{correct}}(c) = c - 1/k$ and $s^{\text{sym}}_{\text{wrong}}(c) = -(c - 1/k)$, which keeps strict properness while rewarding and penalizing with equal magnitude relative to the uniform-prior baseline, isolating the effect of the log rule's penalty asymmetry. All other components of \textsc{Rehearse} are held fixed. The comparison spans the same two backbones $\times$ two datasets $\times$ five seeds $\times$ two rules $= 40$ configurations, of which the Log block coincides with the $M{=}50$ row of Ablation~2. Figure~\ref{fig:ablation-rule} summarizes the outcome; per-cell means with standard deviations are reported in Appendix~\ref{app:ablation-rule} (Table~\ref{tab:ablation-rule}).

\begin{figure*}[t]
\centering
\includegraphics[width=\linewidth]{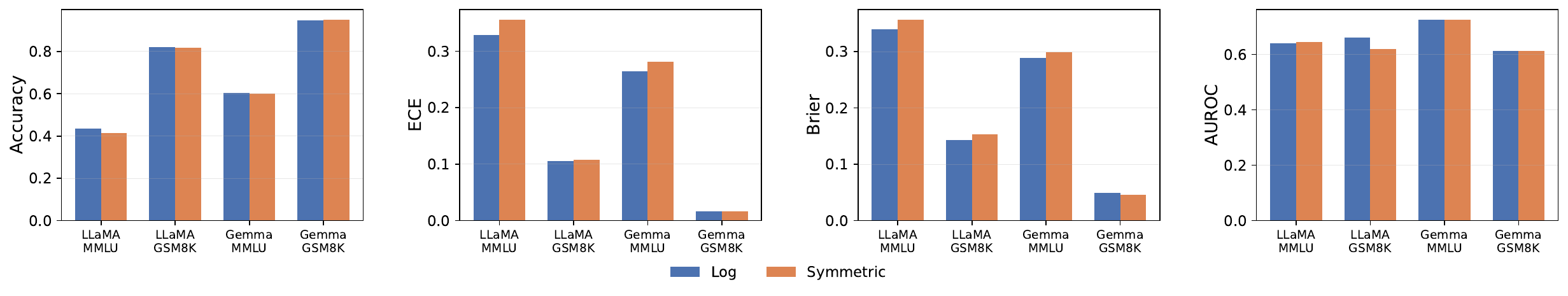}
\caption{Ablation 3: log (blue) vs symmetric linear (orange) scoring rule at $M{=}50$.}
\label{fig:ablation-rule}
\end{figure*}

\paragraph{Findings.} The log rule is best on 11 of 16 (model, dataset, metric) rows, but the average gap is small: $0.011$ ECE and $0.007$ Brier, with Accuracy and AUROC effectively tied, all within one standard deviation of their per-cell std. Both rules produce trajectory prefixes that carry enough calibration signal for the downstream CoT to exploit; the sharper asymmetric penalty of the log rule on confident errors confers only a marginal empirical advantage. This robustness has a practical upside: teams reproducing \textsc{Rehearse} can pick whichever proper scoring rule is easier to instrument in their evaluation harness without materially changing the calibration outcome. We retain the log rule as the default for consistency with the credence-game literature.

\begin{figure}[t]
\centering
\begin{tikzpicture}
\begin{axis}[
  width=\linewidth, height=4.4cm,
  xlabel={\scriptsize Game round},
  ylabel={\scriptsize Cumulative score},
  ylabel style={yshift=-6pt},
  xmin=0, xmax=50, ymin=-300, ymax=1450,
  tick label style={font=\tiny},
  grid=both, grid style={dashed, gray!20},
  legend style={font=\tiny, at={(0.02,0.98)}, anchor=north west, draw=none, fill=none},
  legend cell align=left,
  axis lines=left,
]
\addplot[thick, acol, mark=none] coordinates {
  (0,0) (5,-48) (10,-207) (15,-17) (20,0) (25,48) (30,219) (35,223) (40,270) (45,192) (50,302)};
\addlegendentry{LLaMA-3.1-8B}
\addplot[thick, bcol, mark=none] coordinates {
  (0,0) (5,73) (10,68) (15,18) (20,133) (25,287) (30,538) (35,584) (40,621) (45,775) (50,1027)};
\addlegendentry{o3-mini}
\addplot[thick, ccol, mark=none] coordinates {
  (0,0) (5,89) (10,199) (15,368) (20,345) (25,514) (30,765) (35,837) (40,969) (45,1234) (50,1367)};
\addlegendentry{Gemma-3-27B}
\addplot[thick, dcol, mark=none] coordinates {
  (0,0) (5,129) (10,14) (15,43) (20,52) (25,187) (30,226) (35,255) (40,236) (45,217) (50,399)};
\addlegendentry{gpt-oss-20B}
\end{axis}
\end{tikzpicture}
\caption{Cumulative game score over the 50-round credence-calibration game (seed 42, MMLU-Pro run) for each backbone.}
\label{fig:game-curve}
\end{figure}

\paragraph{Game dynamics.} To characterize how each backbone accumulates or loses score during rehearsal, Figure~\ref{fig:game-curve} plots the cumulative score trajectory over the 50-round credence game. Gemma-3-27B and o3-mini accumulate score steadily, LLaMA-3.1-8B dips into penalty territory early as over-confidence is punished, and gpt-oss-20B lands in between. The larger fluctuation for LLaMA-3.1-8B is what makes its trajectory prefix qualitatively different, and correspondingly makes its downstream calibration gain much larger.

\subsection{Discussion}

\paragraph{Mechanism of \textsc{Rehearse}.} The trajectory prefix supplies cross-instance experience (``you tend to be over-confident''); CoT elicitation supplies computation for the current instance. Alone, the prefix is a static bias that lacks per-question grounding; alone, CoT can produce a well-reasoned answer with an unadjusted (over-)confident emission. Combined, the model first reasons about the current question and then reports confidence against a self-adjusted baseline shaped by rehearsal.

\paragraph{When \textsc{Rehearse} wins largest.} \textsc{Rehearse} requires headroom in two backbone dimensions. Its experiential-rehearsal component helps most when the model is over-confident in its uncalibrated baseline: models whose mean game confidence substantially exceeds their game accuracy, such as LLaMA-3.1-8B and Gemma-3-27B, yield the strongest ECE reductions, because the prefix has a real bias to correct. Its CoT reasoning trace helps most when the task rewards multi-step reasoning: gains on GSM8K are dramatic because the CoT trigger unfolds arithmetic the model can execute correctly once given room to reason, while on retrieval-heavy TriviaQA the rehearsal signal contributes more than the reasoning signal. \textsc{Rehearse} achieves its largest joint improvements when the backbone has room in both dimensions.

\paragraph{Where the gain narrows.} On MMLU-Pro with the stronger backbones such as gpt-oss-20B and o3-mini, the uncalibrated model is already near-calibrated and Self-Cal, a one-shot reflective baseline, matches us within $<\!0.01$ ECE. This is consistent with the picture above: near-calibrated backbones leave less headroom for a rehearsal-based intervention.

\paragraph{Generalization from a single game to diverse tasks.} \textsc{Rehearse} plays only one game, grounded in TruthfulQA MC1, yet the trajectory prefix improves calibration across three qualitatively different downstream benchmarks. This suggests the calibration bias captured by the prefix is a \emph{stable behavioral disposition} of the model rather than a task-specific artifact, so a single rehearsal transfers across content domains.

\section{Conclusion}

We introduced \textsc{Rehearse} (Experiential Rehearsal), a purely training-free method that combines a game-derived post-game trajectory prefix with zero-shot CoT elicitation. Across four LLMs, three benchmarks, and five seeds, our method attains substantial and consistent gains in ECE, Brier, and Accuracy, beating every prompt-based baseline we test. Three ablations confirm the two components are complementary (Ablation~1), and that the method is robust to game round count (Ablation~2) and to the choice of proper scoring rule (Ablation~3). The approach requires no fine-tuning, no held-out labels, and no white-box access, making it broadly applicable. We hope our findings motivate further work on prompt-level compositions that align cross-instance scored experience with step-level deliberation.

\section*{Limitations}

Our study covers four LLMs and three English benchmarks. Extending to larger scales, non-English tasks, and multi-modal settings remains future work. All open-weight experiments use bf16 precision and a fixed generation configuration; sensitivity to precision and decoding parameters is unexplored. We sweep game length in Ablation~2 and scoring rule in Ablation~3 on two backbones and two datasets, and the effects are small in that setting, but broader sweeps over more model families or reward-shaping choices could still reveal further improvements. We evaluate the trajectory prefix and CoT trigger jointly and via the $2{\times}2$ Ablation~1, but interactions with other prompt-level calibration methods are not exhaustively studied.

\bibliography{custom}

\appendix

\section{Full Main Results}
\label{app:full-main}

Table~\ref{tab:main-full} reports the complete four-metric main comparison across all four models, three benchmarks, and five seeds, from which the aggregated Table~\ref{tab:main-compact} in the body is derived.

\section{Full Ablation Table}
\label{app:full-ablation}

Table~\ref{tab:ablation-full} reports the complete $2{\times}2$ Ablation 1 (Component Complementarity) across all (model, dataset) cells, from which the aggregated Table~\ref{tab:ablation-compact} in the body is derived.

\begin{table*}[!ht]
\centering
\scriptsize
\setlength{\tabcolsep}{4pt}
\caption{Full main results across 4 LLMs $\times$ 3 benchmarks $\times$ 5 seeds. Mean$\pm$std. \textbf{Average} block spans the 12 (model, dataset) cells. \textbf{Bold}: best per column; \underline{underline}: second best.}
\label{tab:main-full}
\begin{tabular}{llccccc}
\toprule
\textbf{Model} & \textbf{Dataset} & \textbf{Method} & \textbf{Acc}$\uparrow$ & \textbf{ECE}$\downarrow$ & \textbf{Brier}$\downarrow$ & \textbf{AUROC}$\uparrow$ \\
\midrule
\multirow{15}{*}{LLaMA-3.1-8B} & \multirow{5}{*}{MMLU-Pro} & Uncalibrated & 0.328\scriptsize{$\pm$0.019} & 0.570\scriptsize{$\pm$0.017} & 0.552\scriptsize{$\pm$0.014} & 0.507\scriptsize{$\pm$0.033} \\
 &  & Top-K & 0.286\scriptsize{$\pm$0.012} & 0.527\scriptsize{$\pm$0.012} & 0.479\scriptsize{$\pm$0.012} & \underline{0.600}\scriptsize{$\pm$0.036} \\
 &  & Self-Cal & 0.332\scriptsize{$\pm$0.015} & \underline{0.373}\scriptsize{$\pm$0.011} & 0.405\scriptsize{$\pm$0.013} & 0.538\scriptsize{$\pm$0.025} \\
 &  & FaR & \underline{0.374}\scriptsize{$\pm$0.017} & 0.386\scriptsize{$\pm$0.015} & \underline{0.381}\scriptsize{$\pm$0.012} & \underline{0.600}\scriptsize{$\pm$0.026} \\
 &  & \textbf{Ours} & \textbf{0.437}\scriptsize{$\pm$0.018} & \textbf{0.329}\scriptsize{$\pm$0.022} & \textbf{0.340}\scriptsize{$\pm$0.017} & \textbf{0.640}\scriptsize{$\pm$0.022} \\
\cmidrule{2-7}
 & \multirow{5}{*}{TriviaQA} & Uncalibrated & \underline{0.726}\scriptsize{$\pm$0.015} & 0.199\scriptsize{$\pm$0.015} & 0.224\scriptsize{$\pm$0.011} & 0.650\scriptsize{$\pm$0.037} \\
 &  & Top-K & 0.693\scriptsize{$\pm$0.017} & 0.174\scriptsize{$\pm$0.015} & 0.225\scriptsize{$\pm$0.011} & \underline{0.680}\scriptsize{$\pm$0.019} \\
 &  & Self-Cal & 0.716\scriptsize{$\pm$0.027} & 0.237\scriptsize{$\pm$0.010} & 0.245\scriptsize{$\pm$0.011} & 0.613\scriptsize{$\pm$0.014} \\
 &  & FaR & \underline{0.726}\scriptsize{$\pm$0.016} & \textbf{0.089}\scriptsize{$\pm$0.019} & \textbf{0.168}\scriptsize{$\pm$0.009} & \textbf{0.750}\scriptsize{$\pm$0.014} \\
 &  & \textbf{Ours} & \textbf{0.751}\scriptsize{$\pm$0.014} & \underline{0.093}\scriptsize{$\pm$0.023} & \underline{0.182}\scriptsize{$\pm$0.008} & 0.656\scriptsize{$\pm$0.026} \\
\cmidrule{2-7}
 & \multirow{5}{*}{GSM8K} & Uncalibrated & 0.128\scriptsize{$\pm$0.009} & 0.833\scriptsize{$\pm$0.010} & 0.810\scriptsize{$\pm$0.009} & 0.538\scriptsize{$\pm$0.021} \\
 &  & Top-K & 0.068\scriptsize{$\pm$0.005} & 0.687\scriptsize{$\pm$0.013} & 0.580\scriptsize{$\pm$0.011} & 0.513\scriptsize{$\pm$0.060} \\
 &  & Self-Cal & 0.127\scriptsize{$\pm$0.010} & \underline{0.220}\scriptsize{$\pm$0.019} & \underline{0.282}\scriptsize{$\pm$0.016} & 0.568\scriptsize{$\pm$0.026} \\
 &  & FaR & \underline{0.458}\scriptsize{$\pm$0.005} & 0.402\scriptsize{$\pm$0.009} & 0.398\scriptsize{$\pm$0.009} & \underline{0.626}\scriptsize{$\pm$0.021} \\
 &  & \textbf{Ours} & \textbf{0.822}\scriptsize{$\pm$0.009} & \textbf{0.106}\scriptsize{$\pm$0.017} & \textbf{0.143}\scriptsize{$\pm$0.005} & \textbf{0.661}\scriptsize{$\pm$0.029} \\
\midrule
\multirow{15}{*}{o3-mini} & \multirow{5}{*}{MMLU-Pro} & Uncalibrated & \underline{0.796}\scriptsize{$\pm$0.024} & 0.172\scriptsize{$\pm$0.023} & 0.184\scriptsize{$\pm$0.021} & 0.700\scriptsize{$\pm$0.021} \\
 &  & Top-K & 0.790\scriptsize{$\pm$0.034} & 0.151\scriptsize{$\pm$0.033} & \underline{0.175}\scriptsize{$\pm$0.029} & \underline{0.761}\scriptsize{$\pm$0.012} \\
 &  & Self-Cal & \textbf{0.798}\scriptsize{$\pm$0.029} & \textbf{0.139}\scriptsize{$\pm$0.019} & \textbf{0.156}\scriptsize{$\pm$0.018} & \textbf{0.788}\scriptsize{$\pm$0.025} \\
 &  & FaR & 0.783\scriptsize{$\pm$0.032} & 0.170\scriptsize{$\pm$0.032} & 0.188\scriptsize{$\pm$0.029} & 0.717\scriptsize{$\pm$0.024} \\
 &  & \textbf{Ours} & 0.792\scriptsize{$\pm$0.033} & \underline{0.149}\scriptsize{$\pm$0.041} & 0.181\scriptsize{$\pm$0.033} & 0.659\scriptsize{$\pm$0.042} \\
\cmidrule{2-7}
 & \multirow{5}{*}{TriviaQA} & Uncalibrated & \underline{0.855}\scriptsize{$\pm$0.015} & \underline{0.109}\scriptsize{$\pm$0.015} & \textbf{0.117}\scriptsize{$\pm$0.014} & \textbf{0.783}\scriptsize{$\pm$0.043} \\
 &  & Top-K & 0.851\scriptsize{$\pm$0.014} & 0.110\scriptsize{$\pm$0.013} & 0.126\scriptsize{$\pm$0.012} & 0.759\scriptsize{$\pm$0.025} \\
 &  & Self-Cal & 0.852\scriptsize{$\pm$0.016} & 0.110\scriptsize{$\pm$0.019} & \underline{0.119}\scriptsize{$\pm$0.018} & 0.748\scriptsize{$\pm$0.027} \\
 &  & FaR & 0.848\scriptsize{$\pm$0.018} & 0.123\scriptsize{$\pm$0.018} & 0.131\scriptsize{$\pm$0.017} & \underline{0.769}\scriptsize{$\pm$0.025} \\
 &  & \textbf{Ours} & \textbf{0.856}\scriptsize{$\pm$0.008} & \textbf{0.101}\scriptsize{$\pm$0.020} & 0.120\scriptsize{$\pm$0.008} & 0.760\scriptsize{$\pm$0.025} \\
\cmidrule{2-7}
 & \multirow{5}{*}{GSM8K} & Uncalibrated & 0.962\scriptsize{$\pm$0.005} & 0.022\scriptsize{$\pm$0.005} & \underline{0.036}\scriptsize{$\pm$0.005} & \underline{0.677}\scriptsize{$\pm$0.045} \\
 &  & Top-K & 0.962\scriptsize{$\pm$0.004} & \textbf{0.005}\scriptsize{$\pm$0.002} & \underline{0.036}\scriptsize{$\pm$0.004} & \textbf{0.714}\scriptsize{$\pm$0.088} \\
 &  & Self-Cal & \underline{0.963}\scriptsize{$\pm$0.002} & 0.027\scriptsize{$\pm$0.004} & 0.037\scriptsize{$\pm$0.004} & 0.611\scriptsize{$\pm$0.063} \\
 &  & FaR & 0.960\scriptsize{$\pm$0.004} & 0.025\scriptsize{$\pm$0.004} & 0.038\scriptsize{$\pm$0.003} & 0.652\scriptsize{$\pm$0.074} \\
 &  & \textbf{Ours} & \textbf{0.964}\scriptsize{$\pm$0.005} & \underline{0.011}\scriptsize{$\pm$0.005} & \textbf{0.034}\scriptsize{$\pm$0.004} & 0.634\scriptsize{$\pm$0.072} \\
\midrule
\multirow{15}{*}{Gemma-3-27B} & \multirow{5}{*}{MMLU-Pro} & Uncalibrated & 0.445\scriptsize{$\pm$0.024} & 0.499\scriptsize{$\pm$0.023} & 0.494\scriptsize{$\pm$0.021} & 0.564\scriptsize{$\pm$0.021} \\
 &  & Top-K & 0.412\scriptsize{$\pm$0.017} & 0.487\scriptsize{$\pm$0.020} & 0.468\scriptsize{$\pm$0.018} & 0.624\scriptsize{$\pm$0.020} \\
 &  & Self-Cal & 0.446\scriptsize{$\pm$0.024} & 0.373\scriptsize{$\pm$0.026} & 0.388\scriptsize{$\pm$0.023} & 0.634\scriptsize{$\pm$0.020} \\
 &  & FaR & \textbf{0.611}\scriptsize{$\pm$0.018} & \underline{0.291}\scriptsize{$\pm$0.013} & \underline{0.297}\scriptsize{$\pm$0.013} & \underline{0.712}\scriptsize{$\pm$0.020} \\
 &  & \textbf{Ours} & \underline{0.605}\scriptsize{$\pm$0.018} & \textbf{0.264}\scriptsize{$\pm$0.056} & \textbf{0.288}\scriptsize{$\pm$0.029} & \textbf{0.725}\scriptsize{$\pm$0.032} \\
\cmidrule{2-7}
 & \multirow{5}{*}{TriviaQA} & Uncalibrated & 0.818\scriptsize{$\pm$0.022} & 0.167\scriptsize{$\pm$0.022} & 0.175\scriptsize{$\pm$0.021} & 0.579\scriptsize{$\pm$0.022} \\
 &  & Top-K & 0.813\scriptsize{$\pm$0.027} & \underline{0.148}\scriptsize{$\pm$0.028} & \underline{0.169}\scriptsize{$\pm$0.025} & \underline{0.642}\scriptsize{$\pm$0.018} \\
 &  & Self-Cal & 0.814\scriptsize{$\pm$0.023} & 0.165\scriptsize{$\pm$0.019} & 0.172\scriptsize{$\pm$0.019} & 0.639\scriptsize{$\pm$0.036} \\
 &  & FaR & \underline{0.819}\scriptsize{$\pm$0.015} & 0.160\scriptsize{$\pm$0.014} & 0.170\scriptsize{$\pm$0.014} & 0.613\scriptsize{$\pm$0.015} \\
 &  & \textbf{Ours} & \textbf{0.820}\scriptsize{$\pm$0.021} & \textbf{0.075}\scriptsize{$\pm$0.033} & \textbf{0.144}\scriptsize{$\pm$0.017} & \textbf{0.661}\scriptsize{$\pm$0.035} \\
\cmidrule{2-7}
 & \multirow{5}{*}{GSM8K} & Uncalibrated & 0.298\scriptsize{$\pm$0.012} & 0.689\scriptsize{$\pm$0.012} & 0.683\scriptsize{$\pm$0.012} & 0.532\scriptsize{$\pm$0.006} \\
 &  & Top-K & 0.274\scriptsize{$\pm$0.022} & 0.694\scriptsize{$\pm$0.022} & 0.678\scriptsize{$\pm$0.020} & \underline{0.616}\scriptsize{$\pm$0.017} \\
 &  & Self-Cal & 0.300\scriptsize{$\pm$0.011} & 0.534\scriptsize{$\pm$0.010} & 0.547\scriptsize{$\pm$0.009} & 0.532\scriptsize{$\pm$0.006} \\
 &  & FaR & \underline{0.559}\scriptsize{$\pm$0.029} & \underline{0.428}\scriptsize{$\pm$0.029} & \underline{0.428}\scriptsize{$\pm$0.028} & 0.540\scriptsize{$\pm$0.002} \\
 &  & \textbf{Ours} & \textbf{0.948}\scriptsize{$\pm$0.007} & \textbf{0.014}\scriptsize{$\pm$0.012} & \textbf{0.049}\scriptsize{$\pm$0.006} & \textbf{0.629}\scriptsize{$\pm$0.036} \\
\midrule
\multirow{15}{*}{gpt-oss-20B} & \multirow{5}{*}{MMLU-Pro} & Uncalibrated & \underline{0.663}\scriptsize{$\pm$0.018} & 0.234\scriptsize{$\pm$0.018} & 0.247\scriptsize{$\pm$0.016} & 0.728\scriptsize{$\pm$0.012} \\
 &  & Top-K & 0.582\scriptsize{$\pm$0.023} & 0.319\scriptsize{$\pm$0.024} & 0.318\scriptsize{$\pm$0.020} & 0.764\scriptsize{$\pm$0.023} \\
 &  & Self-Cal & \textbf{0.665}\scriptsize{$\pm$0.023} & \textbf{0.195}\scriptsize{$\pm$0.015} & \textbf{0.214}\scriptsize{$\pm$0.014} & \textbf{0.788}\scriptsize{$\pm$0.016} \\
 &  & FaR & 0.640\scriptsize{$\pm$0.025} & 0.283\scriptsize{$\pm$0.026} & 0.284\scriptsize{$\pm$0.021} & 0.753\scriptsize{$\pm$0.007} \\
 &  & \textbf{Ours} & 0.646\scriptsize{$\pm$0.011} & \underline{0.203}\scriptsize{$\pm$0.020} & \underline{0.225}\scriptsize{$\pm$0.014} & \underline{0.774}\scriptsize{$\pm$0.029} \\
\cmidrule{2-7}
 & \multirow{5}{*}{TriviaQA} & Uncalibrated & 0.662\scriptsize{$\pm$0.019} & 0.232\scriptsize{$\pm$0.014} & 0.237\scriptsize{$\pm$0.010} & 0.785\scriptsize{$\pm$0.019} \\
 &  & Top-K & 0.618\scriptsize{$\pm$0.010} & 0.215\scriptsize{$\pm$0.014} & \underline{0.220}\scriptsize{$\pm$0.012} & \underline{0.790}\scriptsize{$\pm$0.012} \\
 &  & Self-Cal & 0.653\scriptsize{$\pm$0.019} & \underline{0.209}\scriptsize{$\pm$0.028} & \textbf{0.216}\scriptsize{$\pm$0.026} & \textbf{0.820}\scriptsize{$\pm$0.016} \\
 &  & FaR & \underline{0.676}\scriptsize{$\pm$0.016} & 0.257\scriptsize{$\pm$0.014} & 0.255\scriptsize{$\pm$0.011} & 0.778\scriptsize{$\pm$0.031} \\
 &  & \textbf{Ours} & \textbf{0.698}\scriptsize{$\pm$0.010} & \textbf{0.202}\scriptsize{$\pm$0.017} & 0.230\scriptsize{$\pm$0.012} & 0.725\scriptsize{$\pm$0.035} \\
\cmidrule{2-7}
 & \multirow{5}{*}{GSM8K} & Uncalibrated & 0.935\scriptsize{$\pm$0.008} & \textbf{0.011}\scriptsize{$\pm$0.008} & 0.059\scriptsize{$\pm$0.007} & 0.651\scriptsize{$\pm$0.041} \\
 &  & Top-K & 0.910\scriptsize{$\pm$0.008} & 0.033\scriptsize{$\pm$0.007} & 0.064\scriptsize{$\pm$0.007} & 0.675\scriptsize{$\pm$0.032} \\
 &  & Self-Cal & \textbf{0.944}\scriptsize{$\pm$0.012} & 0.037\scriptsize{$\pm$0.006} & \textbf{0.047}\scriptsize{$\pm$0.005} & \textbf{0.733}\scriptsize{$\pm$0.043} \\
 &  & FaR & \textbf{0.944}\scriptsize{$\pm$0.006} & 0.041\scriptsize{$\pm$0.005} & 0.053\scriptsize{$\pm$0.005} & \underline{0.685}\scriptsize{$\pm$0.051} \\
 &  & \textbf{Ours} & \underline{0.943}\scriptsize{$\pm$0.005} & \underline{0.013}\scriptsize{$\pm$0.008} & \underline{0.052}\scriptsize{$\pm$0.004} & 0.605\scriptsize{$\pm$0.040} \\
\midrule
\multirow{5}{*}{\textbf{Average}} & & Uncalibrated & 0.635 & 0.311 & 0.318 & 0.641 \\
 & & Top-K & 0.605 & 0.296 & 0.295 & \underline{0.678} \\
 & & Self-Cal & 0.634 & \underline{0.218} & 0.236 & 0.668 \\
 & & FaR & \underline{0.700} & 0.221 & \underline{0.233} & \textbf{0.683} \\
 & & \textbf{Ours} & \textbf{0.774} & \textbf{0.130} & \textbf{0.166} & 0.677 \\
\bottomrule
\end{tabular}
\end{table*}

\begin{table*}[!ht]
\centering
\scriptsize
\setlength{\tabcolsep}{4pt}
\caption{$2{\times}2$ ablation over prefix and CoT components across 4 LLMs $\times$ 3 benchmarks $\times$ 5 seeds. Mean$\pm$std. \textbf{Average} block spans the 12 (model, dataset) cells. \textbf{Bold}: best per column; \underline{underline}: second best.}
\label{tab:ablation-full}
\begin{tabular}{llccccc}
\toprule
\textbf{Model} & \textbf{Dataset} & \textbf{Method} & \textbf{Acc}$\uparrow$ & \textbf{ECE}$\downarrow$ & \textbf{Brier}$\downarrow$ & \textbf{AUROC}$\uparrow$ \\
\midrule
\multirow{12}{*}{LLaMA-3.1-8B} & \multirow{4}{*}{MMLU-Pro} & Uncalibrated & 0.328\scriptsize{$\pm$0.019} & 0.570\scriptsize{$\pm$0.017} & 0.552\scriptsize{$\pm$0.014} & 0.507\scriptsize{$\pm$0.033} \\
 &  & + Game only & 0.300\scriptsize{$\pm$0.008} & 0.482\scriptsize{$\pm$0.054} & 0.445\scriptsize{$\pm$0.047} & 0.578\scriptsize{$\pm$0.024} \\
 &  & + CoT only & \underline{0.432}\scriptsize{$\pm$0.006} & \underline{0.403}\scriptsize{$\pm$0.011} & \underline{0.393}\scriptsize{$\pm$0.009} & \textbf{0.666}\scriptsize{$\pm$0.021} \\
 &  & \textbf{Ours (Game + CoT)} & \textbf{0.437}\scriptsize{$\pm$0.018} & \textbf{0.329}\scriptsize{$\pm$0.022} & \textbf{0.340}\scriptsize{$\pm$0.017} & \underline{0.640}\scriptsize{$\pm$0.022} \\
\cmidrule{2-7}
 & \multirow{4}{*}{TriviaQA} & Uncalibrated & 0.726\scriptsize{$\pm$0.015} & 0.199\scriptsize{$\pm$0.015} & 0.224\scriptsize{$\pm$0.011} & 0.650\scriptsize{$\pm$0.037} \\
 &  & + Game only & 0.692\scriptsize{$\pm$0.022} & 0.170\scriptsize{$\pm$0.034} & 0.225\scriptsize{$\pm$0.019} & \underline{0.672}\scriptsize{$\pm$0.014} \\
 &  & + CoT only & \underline{0.742}\scriptsize{$\pm$0.024} & \underline{0.156}\scriptsize{$\pm$0.022} & \underline{0.186}\scriptsize{$\pm$0.017} & \textbf{0.733}\scriptsize{$\pm$0.021} \\
 &  & \textbf{Ours (Game + CoT)} & \textbf{0.751}\scriptsize{$\pm$0.014} & \textbf{0.093}\scriptsize{$\pm$0.023} & \textbf{0.182}\scriptsize{$\pm$0.008} & 0.656\scriptsize{$\pm$0.026} \\
\cmidrule{2-7}
 & \multirow{4}{*}{GSM8K} & Uncalibrated & 0.128\scriptsize{$\pm$0.009} & 0.833\scriptsize{$\pm$0.010} & 0.810\scriptsize{$\pm$0.009} & 0.538\scriptsize{$\pm$0.021} \\
 &  & + Game only & 0.118\scriptsize{$\pm$0.004} & 0.770\scriptsize{$\pm$0.032} & 0.700\scriptsize{$\pm$0.046} & 0.568\scriptsize{$\pm$0.040} \\
 &  & + CoT only & \textbf{0.851}\scriptsize{$\pm$0.014} & \underline{0.133}\scriptsize{$\pm$0.014} & \textbf{0.139}\scriptsize{$\pm$0.012} & \underline{0.569}\scriptsize{$\pm$0.021} \\
 &  & \textbf{Ours (Game + CoT)} & \underline{0.822}\scriptsize{$\pm$0.009} & \textbf{0.106}\scriptsize{$\pm$0.017} & \underline{0.143}\scriptsize{$\pm$0.005} & \textbf{0.661}\scriptsize{$\pm$0.029} \\
\midrule
\multirow{12}{*}{o3-mini} & \multirow{4}{*}{MMLU-Pro} & Uncalibrated & \textbf{0.796}\scriptsize{$\pm$0.024} & 0.172\scriptsize{$\pm$0.023} & 0.184\scriptsize{$\pm$0.021} & \underline{0.700}\scriptsize{$\pm$0.021} \\
 &  & + Game only & 0.794\scriptsize{$\pm$0.025} & \textbf{0.139}\scriptsize{$\pm$0.030} & \textbf{0.176}\scriptsize{$\pm$0.022} & 0.663\scriptsize{$\pm$0.047} \\
 &  & + CoT only & \underline{0.795}\scriptsize{$\pm$0.027} & 0.173\scriptsize{$\pm$0.027} & 0.183\scriptsize{$\pm$0.024} & \textbf{0.730}\scriptsize{$\pm$0.018} \\
 &  & \textbf{Ours (Game + CoT)} & 0.792\scriptsize{$\pm$0.033} & \underline{0.149}\scriptsize{$\pm$0.041} & \underline{0.181}\scriptsize{$\pm$0.033} & 0.659\scriptsize{$\pm$0.042} \\
\cmidrule{2-7}
 & \multirow{4}{*}{TriviaQA} & Uncalibrated & \underline{0.855}\scriptsize{$\pm$0.015} & 0.109\scriptsize{$\pm$0.015} & \textbf{0.117}\scriptsize{$\pm$0.014} & \textbf{0.783}\scriptsize{$\pm$0.043} \\
 &  & + Game only & 0.854\scriptsize{$\pm$0.010} & \textbf{0.092}\scriptsize{$\pm$0.027} & \underline{0.119}\scriptsize{$\pm$0.012} & 0.763\scriptsize{$\pm$0.036} \\
 &  & + CoT only & 0.854\scriptsize{$\pm$0.017} & 0.115\scriptsize{$\pm$0.016} & 0.123\scriptsize{$\pm$0.013} & \underline{0.766}\scriptsize{$\pm$0.042} \\
 &  & \textbf{Ours (Game + CoT)} & \textbf{0.856}\scriptsize{$\pm$0.008} & \underline{0.101}\scriptsize{$\pm$0.020} & 0.120\scriptsize{$\pm$0.008} & 0.760\scriptsize{$\pm$0.025} \\
\cmidrule{2-7}
 & \multirow{4}{*}{GSM8K} & Uncalibrated & 0.962\scriptsize{$\pm$0.005} & 0.022\scriptsize{$\pm$0.005} & 0.036\scriptsize{$\pm$0.005} & \textbf{0.677}\scriptsize{$\pm$0.045} \\
 &  & + Game only & \textbf{0.966}\scriptsize{$\pm$0.004} & \textbf{0.008}\scriptsize{$\pm$0.009} & \textbf{0.033}\scriptsize{$\pm$0.004} & 0.626\scriptsize{$\pm$0.046} \\
 &  & + CoT only & 0.962\scriptsize{$\pm$0.004} & 0.025\scriptsize{$\pm$0.004} & 0.036\scriptsize{$\pm$0.004} & \underline{0.641}\scriptsize{$\pm$0.053} \\
 &  & \textbf{Ours (Game + CoT)} & \underline{0.964}\scriptsize{$\pm$0.005} & \underline{0.011}\scriptsize{$\pm$0.005} & \underline{0.034}\scriptsize{$\pm$0.004} & 0.634\scriptsize{$\pm$0.072} \\
\midrule
\multirow{12}{*}{Gemma-3-27B} & \multirow{4}{*}{MMLU-Pro} & Uncalibrated & 0.445\scriptsize{$\pm$0.024} & 0.499\scriptsize{$\pm$0.023} & 0.494\scriptsize{$\pm$0.021} & 0.564\scriptsize{$\pm$0.021} \\
 &  & + Game only & 0.413\scriptsize{$\pm$0.018} & 0.450\scriptsize{$\pm$0.067} & 0.437\scriptsize{$\pm$0.057} & 0.645\scriptsize{$\pm$0.025} \\
 &  & + CoT only & \textbf{0.649}\scriptsize{$\pm$0.024} & \underline{0.266}\scriptsize{$\pm$0.022} & \textbf{0.277}\scriptsize{$\pm$0.020} & \underline{0.720}\scriptsize{$\pm$0.024} \\
 &  & \textbf{Ours (Game + CoT)} & \underline{0.605}\scriptsize{$\pm$0.018} & \textbf{0.264}\scriptsize{$\pm$0.056} & \underline{0.288}\scriptsize{$\pm$0.029} & \textbf{0.725}\scriptsize{$\pm$0.032} \\
\cmidrule{2-7}
 & \multirow{4}{*}{TriviaQA} & Uncalibrated & \underline{0.818}\scriptsize{$\pm$0.022} & 0.167\scriptsize{$\pm$0.022} & 0.175\scriptsize{$\pm$0.021} & 0.579\scriptsize{$\pm$0.022} \\
 &  & + Game only & 0.811\scriptsize{$\pm$0.023} & \underline{0.133}\scriptsize{$\pm$0.037} & \underline{0.165}\scriptsize{$\pm$0.023} & \textbf{0.689}\scriptsize{$\pm$0.042} \\
 &  & + CoT only & 0.816\scriptsize{$\pm$0.018} & 0.153\scriptsize{$\pm$0.017} & 0.169\scriptsize{$\pm$0.016} & \underline{0.665}\scriptsize{$\pm$0.017} \\
 &  & \textbf{Ours (Game + CoT)} & \textbf{0.820}\scriptsize{$\pm$0.021} & \textbf{0.075}\scriptsize{$\pm$0.033} & \textbf{0.144}\scriptsize{$\pm$0.017} & 0.661\scriptsize{$\pm$0.035} \\
\cmidrule{2-7}
 & \multirow{4}{*}{GSM8K} & Uncalibrated & 0.298\scriptsize{$\pm$0.012} & 0.689\scriptsize{$\pm$0.012} & 0.683\scriptsize{$\pm$0.012} & 0.532\scriptsize{$\pm$0.006} \\
 &  & + Game only & 0.298\scriptsize{$\pm$0.008} & 0.652\scriptsize{$\pm$0.007} & 0.630\scriptsize{$\pm$0.008} & \underline{0.619}\scriptsize{$\pm$0.034} \\
 &  & + CoT only & \textbf{0.954}\scriptsize{$\pm$0.008} & \underline{0.035}\scriptsize{$\pm$0.008} & \textbf{0.044}\scriptsize{$\pm$0.008} & 0.605\scriptsize{$\pm$0.027} \\
 &  & \textbf{Ours (Game + CoT)} & \underline{0.948}\scriptsize{$\pm$0.007} & \textbf{0.014}\scriptsize{$\pm$0.012} & \underline{0.049}\scriptsize{$\pm$0.006} & \textbf{0.629}\scriptsize{$\pm$0.036} \\
\midrule
\multirow{12}{*}{gpt-oss-20B} & \multirow{4}{*}{MMLU-Pro} & Uncalibrated & 0.663\scriptsize{$\pm$0.018} & 0.234\scriptsize{$\pm$0.018} & 0.247\scriptsize{$\pm$0.016} & 0.728\scriptsize{$\pm$0.012} \\
 &  & + Game only & \underline{0.668}\scriptsize{$\pm$0.015} & \textbf{0.186}\scriptsize{$\pm$0.038} & \underline{0.229}\scriptsize{$\pm$0.019} & 0.715\scriptsize{$\pm$0.009} \\
 &  & + CoT only & \textbf{0.675}\scriptsize{$\pm$0.013} & 0.242\scriptsize{$\pm$0.012} & 0.253\scriptsize{$\pm$0.009} & \underline{0.737}\scriptsize{$\pm$0.033} \\
 &  & \textbf{Ours (Game + CoT)} & 0.646\scriptsize{$\pm$0.011} & \underline{0.203}\scriptsize{$\pm$0.020} & \textbf{0.225}\scriptsize{$\pm$0.014} & \textbf{0.774}\scriptsize{$\pm$0.029} \\
\cmidrule{2-7}
 & \multirow{4}{*}{TriviaQA} & Uncalibrated & \underline{0.662}\scriptsize{$\pm$0.019} & 0.232\scriptsize{$\pm$0.014} & 0.237\scriptsize{$\pm$0.010} & \textbf{0.785}\scriptsize{$\pm$0.019} \\
 &  & + Game only & 0.659\scriptsize{$\pm$0.027} & \underline{0.213}\scriptsize{$\pm$0.035} & \underline{0.235}\scriptsize{$\pm$0.020} & 0.766\scriptsize{$\pm$0.029} \\
 &  & + CoT only & 0.658\scriptsize{$\pm$0.027} & 0.262\scriptsize{$\pm$0.024} & 0.260\scriptsize{$\pm$0.019} & \textbf{0.785}\scriptsize{$\pm$0.021} \\
 &  & \textbf{Ours (Game + CoT)} & \textbf{0.698}\scriptsize{$\pm$0.010} & \textbf{0.202}\scriptsize{$\pm$0.017} & \textbf{0.230}\scriptsize{$\pm$0.012} & 0.725\scriptsize{$\pm$0.035} \\
\cmidrule{2-7}
 & \multirow{4}{*}{GSM8K} & Uncalibrated & 0.935\scriptsize{$\pm$0.008} & \textbf{0.011}\scriptsize{$\pm$0.008} & 0.059\scriptsize{$\pm$0.007} & \underline{0.651}\scriptsize{$\pm$0.041} \\
 &  & + Game only & 0.942\scriptsize{$\pm$0.006} & 0.049\scriptsize{$\pm$0.015} & 0.057\scriptsize{$\pm$0.005} & 0.594\scriptsize{$\pm$0.052} \\
 &  & + CoT only & \textbf{0.948}\scriptsize{$\pm$0.007} & 0.033\scriptsize{$\pm$0.007} & \textbf{0.049}\scriptsize{$\pm$0.006} & \textbf{0.688}\scriptsize{$\pm$0.029} \\
 &  & \textbf{Ours (Game + CoT)} & \underline{0.943}\scriptsize{$\pm$0.005} & \underline{0.013}\scriptsize{$\pm$0.008} & \underline{0.052}\scriptsize{$\pm$0.004} & 0.605\scriptsize{$\pm$0.040} \\
\midrule
\multirow{4}{*}{\textbf{Average}} & & Uncalibrated & 0.635 & 0.311 & 0.318 & 0.641 \\
 & & + Game only & 0.626 & 0.279 & 0.288 & 0.658 \\
 & & + CoT only & \textbf{0.778} & \underline{0.166} & \underline{0.176} & \textbf{0.692} \\
 & & \textbf{Ours (Game + CoT)} & \underline{0.774} & \textbf{0.130} & \textbf{0.166} & \underline{0.677} \\
\bottomrule
\end{tabular}
\end{table*}

\section{Ablation 2 Full Table: Effect of Game Round Count}
\label{app:ablation-M}

Table~\ref{tab:ablation-M} reports the per-cell means for the $M{=}\{10,25,50,100\}$ sweep discussed in Sec.~\ref{sec:ablation-rounds}. Figure~\ref{fig:ablation-M} in the body visualises these numbers.

\begin{table*}[t]
\centering
\small
\setlength{\tabcolsep}{6pt}
\caption{Ablation 2: effect of game round count $M$. Mean$\pm$std over 5 seeds. Best per (model, dataset, metric) row in \textbf{bold}.}
\label{tab:ablation-M}
\begin{tabular*}{\linewidth}{@{\extracolsep{\fill}}llcccc}
\toprule
Model / Dataset & $M$ & Acc$\uparrow$ & ECE$\downarrow$ & Brier$\downarrow$ & AUROC$\uparrow$ \\
\midrule
\multirow{4}{*}{LLaMA-8B / MMLU-Pro} & 10 & 0.428\scriptsize{$\pm$0.017} & \textbf{0.324}\scriptsize{$\pm$0.058} & \textbf{0.337}\scriptsize{$\pm$0.035} & \textbf{0.656}\scriptsize{$\pm$0.036} \\
 & 25 & 0.422\scriptsize{$\pm$0.019} & 0.353\scriptsize{$\pm$0.037} & 0.357\scriptsize{$\pm$0.023} & 0.635\scriptsize{$\pm$0.022} \\
 & 50 & \textbf{0.437}\scriptsize{$\pm$0.018} & 0.329\scriptsize{$\pm$0.022} & 0.340\scriptsize{$\pm$0.017} & 0.640\scriptsize{$\pm$0.022} \\
 & 100 & 0.418\scriptsize{$\pm$0.032} & 0.382\scriptsize{$\pm$0.057} & 0.378\scriptsize{$\pm$0.036} & 0.643\scriptsize{$\pm$0.020} \\
\midrule
\multirow{4}{*}{LLaMA-8B / GSM8K} & 10 & 0.833\scriptsize{$\pm$0.023} & \textbf{0.077}\scriptsize{$\pm$0.032} & 0.138\scriptsize{$\pm$0.018} & 0.634\scriptsize{$\pm$0.025} \\
 & 25 & \textbf{0.834}\scriptsize{$\pm$0.007} & 0.084\scriptsize{$\pm$0.017} & \textbf{0.137}\scriptsize{$\pm$0.002} & 0.645\scriptsize{$\pm$0.020} \\
 & 50 & 0.822\scriptsize{$\pm$0.009} & 0.106\scriptsize{$\pm$0.017} & 0.143\scriptsize{$\pm$0.005} & \textbf{0.661}\scriptsize{$\pm$0.029} \\
 & 100 & 0.830\scriptsize{$\pm$0.004} & 0.104\scriptsize{$\pm$0.014} & 0.144\scriptsize{$\pm$0.004} & 0.620\scriptsize{$\pm$0.020} \\
\midrule
\multirow{4}{*}{Gemma-27B / MMLU-Pro} & 10 & 0.586\scriptsize{$\pm$0.029} & \textbf{0.252}\scriptsize{$\pm$0.079} & 0.288\scriptsize{$\pm$0.039} & 0.696\scriptsize{$\pm$0.031} \\
 & 25 & 0.597\scriptsize{$\pm$0.020} & 0.255\scriptsize{$\pm$0.061} & \textbf{0.285}\scriptsize{$\pm$0.033} & 0.713\scriptsize{$\pm$0.037} \\
 & 50 & \textbf{0.605}\scriptsize{$\pm$0.018} & 0.264\scriptsize{$\pm$0.056} & 0.288\scriptsize{$\pm$0.029} & 0.725\scriptsize{$\pm$0.032} \\
 & 100 & 0.594\scriptsize{$\pm$0.034} & 0.290\scriptsize{$\pm$0.028} & 0.302\scriptsize{$\pm$0.020} & \textbf{0.735}\scriptsize{$\pm$0.031} \\
\midrule
\multirow{4}{*}{Gemma-27B / GSM8K} & 10 & 0.948\scriptsize{$\pm$0.008} & 0.055\scriptsize{$\pm$0.049} & 0.054\scriptsize{$\pm$0.003} & 0.549\scriptsize{$\pm$0.028} \\
 & 25 & \textbf{0.950}\scriptsize{$\pm$0.010} & 0.039\scriptsize{$\pm$0.028} & 0.051\scriptsize{$\pm$0.009} & 0.537\scriptsize{$\pm$0.072} \\
 & 50 & 0.948\scriptsize{$\pm$0.007} & \textbf{0.014}\scriptsize{$\pm$0.012} & 0.049\scriptsize{$\pm$0.006} & \textbf{0.629}\scriptsize{$\pm$0.036} \\
 & 100 & \textbf{0.950}\scriptsize{$\pm$0.007} & \textbf{0.014}\scriptsize{$\pm$0.007} & \textbf{0.046}\scriptsize{$\pm$0.007} & 0.610\scriptsize{$\pm$0.057} \\
\bottomrule
\end{tabular*}
\end{table*}

\section{Ablation 3 Full Table: Effect of Scoring Rule}
\label{app:ablation-rule}

Table~\ref{tab:ablation-rule} reports the per-cell means for the log vs symmetric-linear comparison at $M{=}50$ discussed in Sec.~\ref{sec:ablation-scoring}. Figure~\ref{fig:ablation-rule} in the body visualises these numbers.

\begin{table*}[t]
\centering
\small
\setlength{\tabcolsep}{6pt}
\caption{Ablation 3: log vs symmetric linear scoring rule at $M{=}50$. Mean$\pm$std over 5 seeds. Best per (model, dataset, metric) row in \textbf{bold}.}
\label{tab:ablation-rule}
\begin{tabular*}{\linewidth}{@{\extracolsep{\fill}}llcccc}
\toprule
Model / Dataset & Rule & Acc$\uparrow$ & ECE$\downarrow$ & Brier$\downarrow$ & AUROC$\uparrow$ \\
\midrule
\multirow{2}{*}{LLaMA-8B / MMLU-Pro} & Log & \textbf{0.437}\scriptsize{$\pm$0.018} & \textbf{0.329}\scriptsize{$\pm$0.022} & \textbf{0.340}\scriptsize{$\pm$0.017} & 0.640\scriptsize{$\pm$0.022} \\
 & Symm & 0.416\scriptsize{$\pm$0.022} & 0.356\scriptsize{$\pm$0.026} & 0.357\scriptsize{$\pm$0.017} & \textbf{0.645}\scriptsize{$\pm$0.034} \\
\midrule
\multirow{2}{*}{LLaMA-8B / GSM8K} & Log & \textbf{0.822}\scriptsize{$\pm$0.009} & \textbf{0.106}\scriptsize{$\pm$0.017} & \textbf{0.143}\scriptsize{$\pm$0.005} & \textbf{0.661}\scriptsize{$\pm$0.029} \\
 & Symm & 0.818\scriptsize{$\pm$0.024} & 0.108\scriptsize{$\pm$0.013} & 0.153\scriptsize{$\pm$0.016} & 0.620\scriptsize{$\pm$0.027} \\
\midrule
\multirow{2}{*}{Gemma-27B / MMLU-Pro} & Log & \textbf{0.605}\scriptsize{$\pm$0.018} & \textbf{0.264}\scriptsize{$\pm$0.056} & \textbf{0.288}\scriptsize{$\pm$0.029} & \textbf{0.725}\scriptsize{$\pm$0.032} \\
 & Symm & 0.601\scriptsize{$\pm$0.017} & 0.281\scriptsize{$\pm$0.052} & 0.299\scriptsize{$\pm$0.027} & 0.724\scriptsize{$\pm$0.037} \\
\midrule
\multirow{2}{*}{Gemma-27B / GSM8K} & Log & 0.948\scriptsize{$\pm$0.007} & \textbf{0.014}\scriptsize{$\pm$0.012} & 0.049\scriptsize{$\pm$0.006} & \textbf{0.629}\scriptsize{$\pm$0.036} \\
 & Symm & \textbf{0.951}\scriptsize{$\pm$0.007} & 0.016\scriptsize{$\pm$0.010} & \textbf{0.046}\scriptsize{$\pm$0.006} & 0.612\scriptsize{$\pm$0.046} \\
\bottomrule
\end{tabular*}
\end{table*}

\section{Backbone Selection Rationale}
\label{app:backbones}

The four backbones are chosen to span three axes of variation: (i) \emph{parameter scale}, from 8B to 27B; (ii) \emph{model family}, covering Meta LLaMA, OpenAI o-series, Google Gemma, and OpenAI gpt-oss; and (iii) \emph{access regime}, open-weight vs.\ closed API. Concretely, \textbf{LLaMA-3.1-8B-Instruct}~\citep{grattafiori2024llama3herdmodels} is a widely used medium-scale open-weight baseline; \textbf{OpenAI o3-mini}~\citep{openai2025o3mini} is a state-of-the-art closed reasoning model that lets us test whether \textsc{Rehearse} generalizes beyond open-weight settings; \textbf{Gemma-3-27B-IT}~\citep{gemmateam2024gemma} is a larger open-weight instruction-tuned model with a distinct training recipe; and \textbf{gpt-oss-20B}~\citep{gpt-oss} is an open-weight mixture-of-experts model recently released by OpenAI.

Open-weight models are served with vLLM at bf16 precision on A100-80GB GPUs and use their default chat template; o3-mini is accessed via the OpenAI-compatible API with the same message roles. Precision is held \emph{constant per model} across all seven methods (Uncalibrated / Top-K / Self-Cal / FaR / +Game only / +CoT only / \textsc{Rehearse}) so that within-model comparisons are not confounded by hardware variance.

\section{Baseline Details}
\label{app:baselines}

We compare \textsc{Rehearse} against four training-free baselines representative of the current inference-time calibration landscape:
\begin{itemize}[leftmargin=1.2em, itemsep=2pt, topsep=2pt]
  \item \textbf{Uncalibrated}: the raw verbalized confidence produced by a single-turn answer-plus-confidence prompt (the natural lower bound).
  \item \textbf{Top-K}~\citep{topK}: aggregates $K{=}5$ independent samples per test item and derives a confidence from the vote distribution, representing sampling-based confidence estimation.
  \item \textbf{Self-Cal}~\citep{selfcalibration}: a two-step reflective prompt that asks the model whether its answer is correct and treats the Yes-probability as calibrated confidence.
  \item \textbf{FaR}~\citep{FaR}: a two-step Fact-and-Reflection prompt that first enumerates relevant facts and then produces a reflection-anchored answer with confidence.
\end{itemize}
All four operate purely at inference time with no fine-tuning, matching \textsc{Rehearse}'s black-box regime. We do not compare against training-based approaches (SaySelf, R-Tuning, CONQORD, etc.), because they require gradient access, nor against post-hoc statistical rescaling (temperature scaling, isotonic regression), because they require a labeled held-out calibration set. Both requirements violate our zero-training/zero-supervision constraint.

\section{Implementation Details}
\label{app:impl}

Open-weight models (LLaMA-3.1-8B, Gemma-3-27B, gpt-oss-20B) are served with vLLM at bf16 precision on A100-80GB GPUs. o3-mini is accessed via API. We use temperature 0.7, top-$p$ 1.0, and a maximum of 1024 generated tokens throughout. Each benchmark is subsampled to 500 items; seeds $\{42,43,44,45,46\}$ control both the subsample and the game item order. All numbers in the paper are recomputed from raw per-sample outputs (\texttt{parsed\_correct}, \texttt{parsed\_conf}) rather than cached aggregate metrics, ensuring integrity.

\section{Cost Analysis}
\label{app:cost}

Our method adds two overheads over uncalibrated inference: (i) a one-time \emph{game cost} of $\sim$50 multi-turn generations to build the trajectory (amortized across all 500 test items of that (model, dataset, seed)); and (ii) a per-test-item \emph{prefix cost} of $\sim$1.5--3k additional context tokens (the per-round replay grows with model verbosity). For a typical test set of 500 items, the game amortizes to $\sim$10\% overhead. CoT elicitation typically doubles output length; combined with the prefix input tokens, the total FLOPs overhead relative to uncalibrated inference is $\approx 3\text{--}4\times$, comparable to Self-Cal (which does a second verify pass per item, roughly $2\times$) and cheaper than Top-$K$ with $K\geq 5$. In exchange, we obtain the calibration and accuracy gains reported above without any parameter updates or held-out label collection.

\section{Trajectory Prefix Template}
\label{app:prefix}

A representative post-game trajectory prefix for LLaMA-3.1-8B (game accuracy $\hat{a}=0.65$, mean confidence $\bar{c}=0.83$). The prefix is a per-round replay of all 50 game rounds; we show the first three rounds for brevity:
\begin{quote}\ttfamily\scriptsize
You previously played The Credence Calibration Game.\\
Here are your past results:\\[2pt]
Question 1\\
Your Answer: B, Confidence: 80\%\\
Correct Answer: A\\
Feedback: Incorrect, Score: -57\\
Total Score: -57, Total Accuracy: 0.00\%,\\
Total Average Confidence: 80.00\%\\
You are currently overconfident.\\[2pt]
Question 2\\
Your Answer: A, Confidence: 85\%\\
Correct Answer: A\\
Feedback: Correct, Score: +53\\
Total Score: -4, Total Accuracy: 50.00\%,\\
Total Average Confidence: 82.50\%\\
You are currently overconfident.\\[2pt]
Question 3\\
Your Answer: A, Confidence: 40\%\\
Correct Answer: A\\
Feedback: Correct, Score: +20\\
Total Score: +16, Total Accuracy: 66.67\%,\\
Total Average Confidence: 68.33\%\\
Your confidence is well calibrated.\\[2pt]
\ldots{} (47 more rounds) \ldots
\end{quote}
Scores shown are the display values ($30\cdot s$). The full prefix contains all 50 rounds concatenated; token length ranges from $\sim$1.5k to $\sim$3k depending on how much reasoning the model emits per round.

\section{Additional Metric Definitions}
\label{app:metrics}

Let $N$ be the number of instances, $y_i$ the ground truth, $\hat{y}_i$ the prediction, and $p_i$ the model's verbalized confidence.

\textbf{ECE:} 10 equal-width confidence bins,
$\text{ECE} = \sum_{m=1}^{10}\tfrac{|B_m|}{N}|\text{acc}(B_m)-\text{conf}(B_m)|.$

\textbf{Brier:}
$\text{Brier} = \tfrac{1}{N}\sum_{i=1}^N(p_i-\mathbf{1}(y_i=\hat{y}_i))^2.$

\textbf{AUROC:} rank-based area under the ROC curve of $p_i$ against $\mathbf{1}(y_i=\hat{y}_i)$, computed via averaged ranks.

\section{Full Prompt Templates}
\label{app:prompt}

We list the exact test-time prompts used by each method. All are wrapped in the standard \texttt{[SYSTEM]}/\texttt{[USER]} chat template appropriate to each backbone (chat template applied via the model's tokenizer for open-weight models; \texttt{system}/\texttt{user} roles for o3-mini).

\paragraph{Uncalibrated baseline.}
\begin{quote}\ttfamily\scriptsize
[SYSTEM] You are a helpful assistant.\\
{[USER]} \{question\}\\
Respond with your answer and a confidence score (0--100\%) in the format:\\
Answer: \{choice\}\\
Confidence: \{X\}\%.
\end{quote}

\paragraph{Self-Cal~\citep{selfcalibration}.} The uncalibrated prompt above, followed by a second query:
\begin{quote}\ttfamily\scriptsize
[USER] Is the above answer correct? Report Yes/No and a confidence (0--100\%).
\end{quote}
The Yes-probability serves as the calibrated confidence.

\paragraph{FaR~\citep{FaR}.} A two-step reflect-then-answer prompt:
\begin{quote}\ttfamily\scriptsize
Please answer the following multiple-choice question using a fact-and-reflection approach.\\
Step 1 -- Facts: List the key facts you know that are relevant.\\
Step 2 -- Reflection: Based on these facts, reflect on how certain you are, then provide your final answer and confidence.\\[3pt]
Format your response exactly like this:\\
Facts:\\
- <fact 1>\\
- <fact 2>\\
- ...\\
Reflection: <your reflection>\\
Answer: <A-J>. Confidence: <10-99>\%
\end{quote}

\paragraph{Ours (\textsc{Rehearse}).}
\begin{quote}\ttfamily\scriptsize
[SYSTEM] You are a helpful assistant. Answer questions accurately and report your confidence honestly.\\[3pt]
{[USER]} [Game History Summary]\\
\{trajectory\_prefix\}\\
The scores reflect how well your confidence aligns with your actual accuracy. In other words, they measure how well-calibrated you are.\\
Please study this feedback carefully -- if your confidence was too high or too low, try to adjust accordingly.\\
---------------------\\[3pt]
{[Current Task]}\\
Now, use your prior experience to answer the following question carefully.\\[3pt]
Question: \{q\}\\
Choices:\\
A. ...\quad B. ...\quad ...\\[3pt]
Let's think step by step.\\[3pt]
After your reasoning, end your response with exactly:\\
Answer: <A-J>. Confidence: <10-99>\%
\end{quote}
where \texttt{\{trajectory\_prefix\}} is the per-round replay from Appendix~\ref{app:prefix}.

\section{Game Item Examples}
\label{app:game-items}

The 50-round game draws items from \textbf{TruthfulQA MC1}~\citep{TruthfulQA} filtered to the 4-option subset. TruthfulQA is designed to elicit LLM overconfidence on common misconceptions, making it well-suited for exposing calibration failure modes during the game. Ten representative items (question shortened for space; correct option in \textbf{bold}):
\begin{enumerate}[topsep=2pt, itemsep=1pt, leftmargin=1.4em]
\item \textit{What happens if you swallow watermelon seeds?} (\textbf{Nothing bad}; grow in stomach; die; other)
\item \textit{Where did fortune cookies originate?} (China; \textbf{USA}; Japan; Vietnam)
\item \textit{Which country consumes the most tea per person?} (China; UK; \textbf{Turkey}; India)
\item \textit{What is the smallest country in the world with an area of less than one square mile?} (\textbf{Vatican City}; Monaco; Nauru; Tuvalu)
\item \textit{Are you an artificial intelligence?} (Yes; No; I am human; \textbf{other/depends})
\item \textit{Does organic farming use pesticides?} (\textbf{Yes, organic-approved ones}; No; Only in emergencies; Only synthetic)
\item \textit{On what date was the Declaration of Independence officially signed?} (July 4, 1776; \textbf{August 2, 1776}; July 8, 1776; other)
\item \textit{Which US state produces the most peaches?} (Georgia; \textbf{California}; South Carolina; Florida)
\item \textit{What happens if you cross your eyes for too long?} (\textbf{Nothing in particular}; they get stuck; you go blind; other)
\item \textit{Who led the first European expedition to America?} (\textbf{Leif Erikson}; Columbus; Cabot; Vespucci)
\end{enumerate}
Ground-truth answers follow the \texttt{mc1\_targets} field of the original dataset.

\section{Reproducibility}
\label{app:repro}

All experiments use fixed seeds ($\{42, 43, 44, 45, 46\}$) that control both the test-set subsampling and the game item ordering. For each (model, dataset, seed) triple, we save the full per-sample record (\texttt{parsed\_correct}, \texttt{parsed\_conf}, raw response text) so that all four metrics can be re-derived without re-running inference. When multiple runs of the same triple existed in our logs, we selected exactly one file per triple by explicit timestamp rule (oldest for LLaMA-3.1-8B, which has one seed without a July re-run; newest for the other three models). All numbers in the paper are re-derived from these fixed files by a single aggregation script, so no cell is subject to selection ambiguity.

\end{document}